\documentclass[twoside]{article}

\usepackage[accepted]{aistats2023}
\usepackage{graphicx}
\usepackage{tikz}
\usepackage{subfig}
\usepackage[authoryear,sort&compress]{natbib}
\usepackage[hidelinks]{hyperref}
\usepackage{amsfonts}
\usepackage{algpseudocode,algorithm,algorithmicx}
\usepackage{tabularx}
\usepackage[ruled,vlined,linesnumbered,algo2e]{algorithm2e}

\DeclareMathOperator*{\argmax}{arg\,max}

\usetikzlibrary{er,positioning,shapes.geometric}

\begin{document}

%
\runningtitle{Clarity: an improved gradient method for producing quality visual counterfactual explanations}

%

\twocolumn[

\aistatstitle{Clarity: an improved gradient method\\for producing quality visual counterfactual explanations}

\aistatsauthor{Claire Theobald\textsuperscript{1} \And Frédéric Pennerath \textsuperscript{2} \And  Brieuc Conan-Guez \textsuperscript{2} \And Miguel Couceiro\textsuperscript{1} \And Amedeo Napoli\textsuperscript{1} }

\aistatsaddress{\textsuperscript{1}Université de Lorraine, CNRS, LORIA, F-54000 Nancy France \\ \texttt{\{\{claire.theobald, miguel.couceiro, amedeo.napoli\}@loria.fr\}} \\ \and \textsuperscript{2}Université de Lorraine, CentraleSupélec, CNRS, LORIA, F-57000 Metz France \\ \texttt{\{frederic.pennerath@centralesupelec.fr, brieuc.conan-guez@univ-lorraine.fr\}}}]

\begin{abstract}

 Visual counterfactual explanations identify modifications to an image that would change the prediction of a classifier. We propose a set of techniques based on generative models (VAE) and a classifier ensemble directly trained in the latent space, which all together, improve the quality of the gradient required to compute visual counterfactuals. These improvements lead to a novel classification model, \textit{Clarity}, which produces realistic counterfactual explanations over all images. We also present several experiments that give insights on why these techniques lead to better quality results than those in the literature. The explanations produced are competitive with the state-of-the-art and emphasize the importance of selecting a meaningful input space for training.

\end{abstract}

\section{INTRODUCTION}

In many application areas, deep neural networks must not only be powerful predictive tools but they must also produce interpretable results that can be understood and accepted by humans \citep{Sartor20, Callahan17, Singla20}.
One of the techniques contributing to the interpretability of results produced by deep classifiers is the generation of \emph{counterfactuals} \citep{Watcher17, Keane21}:
given a classifier $C$, an input $X$ and its predicted class $y = \argmax(C(X))$ with $C(X)$ the output distribution of the classifier, a \emph{counterfactual} $X'$ of $X$ for a \emph{target class} $y' \neq y$ is an input as close as possible to $X$ but of predicted class $y' = \argmax(C(X'))$.
To do this we add the additional constraint that $X'$ must be a realistic/plausible example of the class $y'$ (i.e., belongs to the collection of examples of the class $y'$), which distinguishes a counterfactual from an \emph{adversarial attack} \citep{Szegedy14}.
A counterfactual thus expresses the minimal modifications that must be made to $X$ so that it is interpreted by the classifier as belonging to the target class $y'$ instead of $y$.

The production of counterfactuals was first applied to low-dimensional tabular data \citep{Watcher17}, where the gap between the counterfactual $X'$ and $X$ can be measured simply by the minimum number of controllable features to modify (e.g. number of renovation works to sell a house).
This problem was then extended to high dimensional spaces such as images to give rise to \emph{visual counterfactual explanations} \citep{Goyal20, Boreiko22, Liu19, Chang19, Sanchez22}.

In this paper we focus on \emph{gradient-based methods} for generating \emph{visual counterfactuals}. Under this term we gather all methods \citep{Schut21, Joshi19, Holtgen21, Liu19} that are based on gradient descent to find the counterfactual image $X'$, typically by searching an image that substantially maximizes the probability $P(Y=y'|X')$ of the target class, starting from $X$. 
This restriction is motivated by the fact that gradient-based methods remain simple and fast (as they can benefit from deep learning frameworks running on GPUs) compared to other more sophisticated counterfactual search techniques, such as those based on vectorial quantification \citep{Looveren21} or sparse changes around an $l_p$-ball, which is sensitive to multiple hyperparameters and adversarial robustness \citep{Boreiko22}.

In this context of gradient-based methods, we propose a new and simpler classification model that is able to produce quality visual counterfactuals, where other gradient-based methods fail to do so, even on an image classification problem as simple as MNIST. 
This new model has been conceived to be explainable \textit{by design}, which can be desirable in areas where explainability is expected. The counterpart is that our method integrates its own classifiers and thus does not allow, unlike other methods \citep{Schut21, Joshi19, Looveren21}, to take any arbitrary classifier as an input. Therefore our method does not serve as a debugging tool to inspect the consistency of the results produced by some external classifier.

We call our model \emph{Clarity}, and it consists in computing the gradient from an \emph{ensemble} of classifiers \emph{trained directly in the latent space of a VAE}. We show that only the combined contributions of the VAE and of a model ensemble enable a stable computation of the gradient (i.e., reduce its variance), ensure the convergence of the gradient descent and, eventually, result in quality counterfactuals. 


To summarize, our main contributions are the following:
\begin{itemize}
    \item a set of techniques to improve the quality of the gradient used to compute counterfactuals;
    \item a classification model \emph{Clarity} based on these techniques to produce quality counterfactual images;
    \item insights based on many experiments that explain why these techniques are essential to obtain quality counterfactuals.
\end{itemize}
The next four sections sequentially describe the difficulties encountered when computing a gradient-based visual counterfactual and how each of these issues can be overcome, leading to our method \emph{Clarity}.

\section{COUNTERFACTUAL EXPLANATIONS IN IMAGE SPACE}
\label{section:image_space}

In this section, we will study how counterfactual explanations behave when applied to deep learning models on images, in the simplest form. Recall that a counterfactual explanation consists in finding the minimal changes to an image $X$, such that the model predicts the new target class: $y'=\argmax(C(X'))$, and $X'=X+\Delta$, where $\Delta$ is the counterfactual perturbation. 

In most methods, counterfactual explanations are obtained by iteratively minimizing the following quantity
\begin{align}
   \mathcal{L}_{CE}(X')=L(C(X'),y') + \lambda \, d(X,X'),
\label{eq:ce_optimization}
\end{align}
where $L$ denotes the cross-entropy loss and $d$ is a distance function.
The first term ensures that the counterfactual explanation is classified as $y'$ by the model, and the second term forces the explanation to not be too far off the original input $X$. Usually $d$ is chosen as the distance function of the L1 or L2 norm. $\lambda$ is a hyperparameter that defines how strongly the counterfactual is penalized from going too far from the original example. One might note that this optimization problem is the same as the one used to generate adversarial examples \citep{Watcher17, Szegedy14}.
This fact raises the problem of ensuring that the counterfactual explanations appear to be \textit{realistic}. 

The notion of realism in counterfactual explanations has been the subject to several discussions in the research community. For tabular data, it can be as easy as checking that the different values make sense, but it does not exclude some cases that can be challenging, especially when the number of attributes and possible values grow. For image data, it can be even more challenging to define what a realistic image looks like. One way to reliably say whether 
a counterfactual image is realistic 
is simply to ask an expert if the new image $X'$ truly belongs to the intended target class $y'$ \citep{Freiesleben22}. If this is not the case, then it means that $X'$ is rather an adversarial example that tricked the classifier into predicting the wrong label.

\begin{figure}[htbp]
\centering
\subfloat[Gradient descent\label{fig:2_cnn}]{%
\includegraphics[scale=0.27, trim=40 13 90 40, clip]{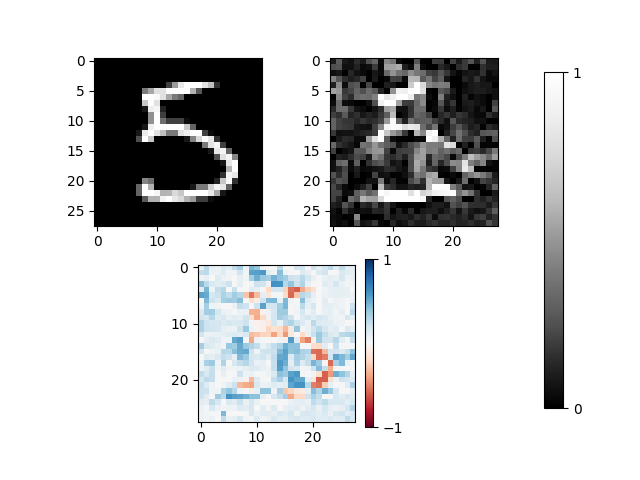}
}\hfil
\subfloat[Schut et al.\label{fig:2_schut}]{%
\includegraphics[scale=0.27, trim=40 13 40 40, clip]{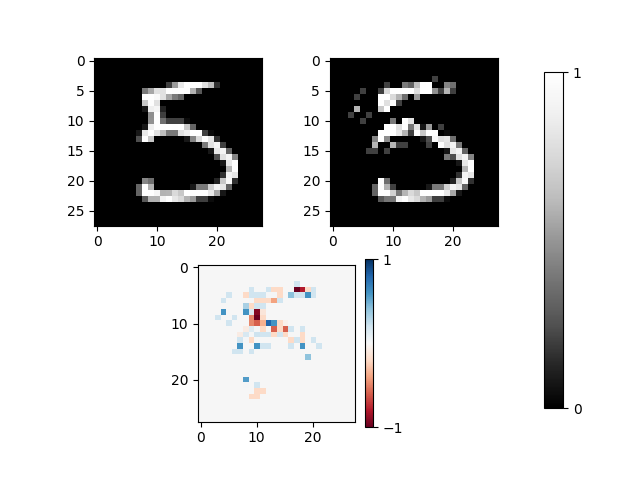}
}\hfil
\vspace{.1in}
\caption{Comparison of counterfactual explanations ($5 \rightarrow 1$) between two image space algorithms. Top left: original image. Top right: Counterfactual explanation. Bottom: Difference between explanation and original image. Schut has failed to converge after 500 iterations.}
\label{fig:counterfactual_image}
\end{figure}

Our first experiment is then to naively optimise \eqref{eq:ce_optimization} by applying a gradient-based algorithm. We train a CNN on MNIST dataset \citep{mnist}, with adversarial training \citep{Goodfellow15}. Adversarial training has been shown to highly improve the quality of counterfactual explanations \citep{Schut21, Boreiko22}. We then generate a counterfactual example by using the Adam optimizer \citep{Kingma15} to solve \eqref{eq:ce_optimization}, similarly to the Fast Gradient Sign Method (FGSM) \citep{Goodfellow15}. We choose L1-norm as the distance, with $\lambda=0.001$ since higher values prevent the counterfactual to converge to the target class. We stop once we reach a target probability $P(Y=y'|X')\geq0.99$ or more than 500 steps. On dozens of attempts, many of them failed to converge to a convincing counterfactual. For instance, Figure \ref{fig:2_cnn} shows the result when trying to generate a counterfactual image of a 1 from an image of a 5. Even with adversarial training, the generated counterfactual is noisy and does not look like a 1 nor like any number at all.

In order to limit the presence of noise in the data, and to obtain sparse counterfactual explanations, Schut et al. \citep{Schut21} propose an algorithm based on the Jacobian-based Saliency Map Attack (JSMA) \citep{Papernot16}. This ensures that the changes are sparse and that the final counterfactual is generated by a minimal amount of changes to the image itself. 

 Their algorithm revolves around a Bayesian Neural Network (BNN) based on an ensemble of models \citep{Laksh17}, in order to implicitly minimize the ``epistemic uncertainty'' of the prediction. \textit{Epistemic uncertainty} represents the uncertainty of the model due to the lack of knowledge in the data distribution. As such, a trained BNN will have higher epistemic uncertainty on unrealistic out-of-distribution samples. They also use \textit{aleatoric uncertainty}, which measures the uncertainty related to the noise in the data itself, to generate unambiguous counterfactuals. Both uncertainties can be retrieved by decomposing the overall \textit{predictive uncertainty} \citep{Kendall17,Depeweg18}. 

In practice, they replace the second term in \eqref{eq:ce_optimization} by the predictive entropy, which measures the overall interpretability of the counterfactual. They show that simply minimizing the expected value of the cross-entropy also minimizes the predictive entropy. By using an ensemble of models $(C_m)_{m=1}^M$, their objective function then becomes
\begin{align}
   \mathcal{L}_{Schut}(X') = \frac{1}{M}\sum_{m=1}^ML(C_m(X'),y').
\label{eq:ce_optimization_schut}
\end{align}
 The results of this algorithm on the same counterfactual example, are shown in Figure \ref{fig:2_schut}. While this algorithm works sometimes on examples where the distance between the image and its counterfactual is minimal, it is not able to create meaningful and realistic counterfactuals for any image. Here, we let the algorithm run for 500 iterations with a step $\delta=0.2$. The result doesn't resemble the intended class, with a final target probability $P(Y=1|X')$ around $10^{-5}$. While the changes are sparse compared to the simple gradient descent algorithm, there is still noise present on the image. The fact that this algorithm is not suited for every couple of starting and target classes was observed in the original paper \citep{Schut21}. We observed this fact on several images of MNIST. More examples are shown in Table~\ref{table:mnist}, where the algorithm either failed to converge or converged to an unrealistic counterfactual.

\section{COUNTERFACTUAL EXPLANATIONS IN A LATENT SPACE}
\label{section:latent_space} 

We now understand that trying to generate realistic counterfactual explanations in a highly dimensional space such as an image space, can be challenging.
A fundamental issue is that we are trying to generate a realistic image representing a high-level concept by modifying low-level data. 

In \citet{Joshi19} the authors propose to restrict the space in which an explanation can be generated. They do this by using a generative model, which produces data samples from a distribution $p(X)$. The most widely used generative models are Generative Adversarial Networks (GANs) \citep{Goodfellow14} and Variational Auto-Encoders (VAEs) \citep{Kingma14}. The generative model chosen in \citet{Joshi19} for most of their experiments was a VAE, although it is also possible on a GAN \citep{Liu19}. In this case, we will use a VAE. Let $q_\theta(z|X)$ be the normal variational posterior which samples a latent variable $z\in \mathcal{Z} \subset \mathbb{R}^d$ from an image $X \in \mathcal{X} \subset \mathbb{R}^k$, with $d \ll k$, thanks to an encoder network  parameterized by $\theta$. Let $z \mapsto \mathcal{G}_{\psi}(z)$ be the generative function, implemented as a decoder network parameterized by $\psi$, that decodes the latent variable $z$ into the image $X$, i.e. $\mathcal{G}_{\psi}(z) \approx X$.

 The counterfactual generation algorithm in \citet{Joshi19} called \emph{REVISE}, aims to optimize the loss function
\begin{align}
   \mathcal{L}_{Revise}(z') = L(C(\mathcal{G}_{\psi}(z')),y') + \lambda \, d(X,\mathcal{G}_{\psi}(z')).
\label{eq:ce_optimization_revise}
\end{align}
Solving this optimization problem can be done using the same gradient-based algorithm, but the optimization is done on $z'$ rather than $X'$. The initial latent embedding is given by the encoder: $z' \sim q_\theta(z|X)$.

\begin{figure}[htbp]
\centering
\subfloat[$6 \rightarrow 3$ (REVISE)\label{fig:6_revise_nobdl}]{%
\includegraphics[scale=0.27, trim=40 13 90 40, clip]{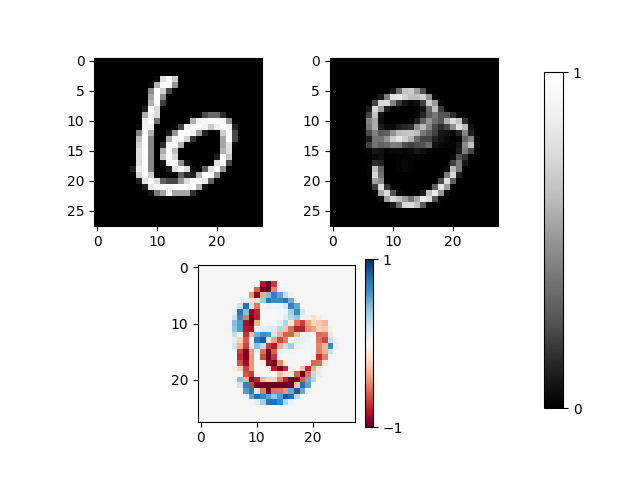}
}\hfil
\subfloat[$5 \rightarrow 1$ (REVISE)\label{fig:5_revise_nobdl}]{%
\includegraphics[scale=0.27, trim=40 13 40 40, clip]{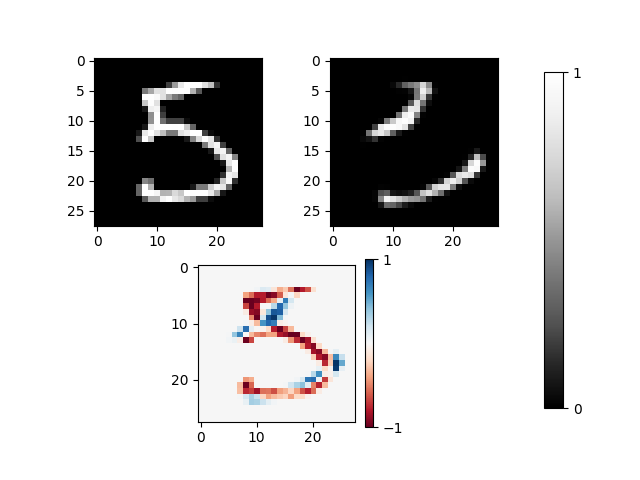}
}\hfil\\
\subfloat[$6 \rightarrow 3$ (REVISE-E)\label{fig:6_revise}]{%
\includegraphics[scale=0.27, trim=40 13 90 40, clip]{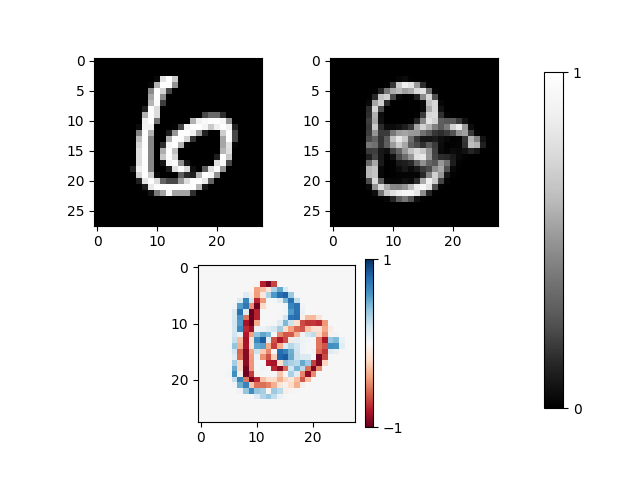}
}\hfil
\subfloat[$5 \rightarrow 1$ (REVISE-E)\label{fig:5_revise}]{%
\includegraphics[scale=0.27, trim=40 13 40 40, clip]{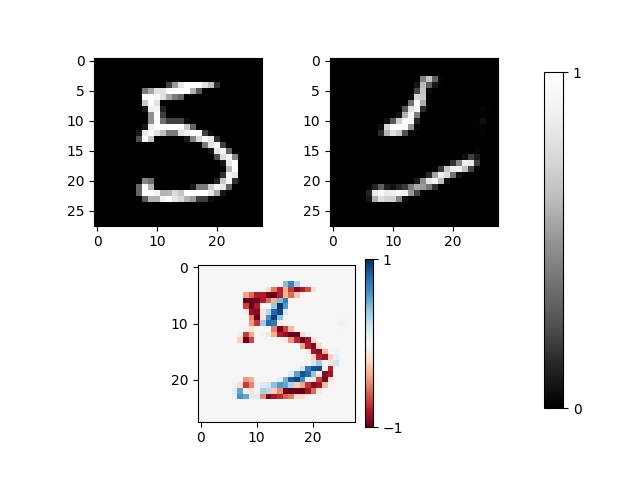}
}\hfil\\
\vspace{.1in}
\caption{Counterfactual explanations using REVISE (\ref{fig:6_revise_nobdl}, \ref{fig:5_revise_nobdl}) and REVISE-E (\ref{fig:6_revise}, \ref{fig:5_revise}).}
\label{fig:counterfactual_revise}
\end{figure}

Figure \ref{fig:6_revise_nobdl} and \ref{fig:5_revise_nobdl} present some counterfactual explanations produced by REVISE.
We use the same CNN as in the image space-based counterfactual algorithms shown in Section \ref{section:image_space}. We use $\lambda=0.001$ with an L1-norm (the algorithm does not converge for higher values of $\lambda$). In both cases, the algorithm converged to an $X'$ with $P(Y=y'|X')\geq0.99$. The counterfactual images obtained by REVISE are generally better and less noisy, but there are still some cases where the counterfactual examples are not realistic. Figure \ref{fig:6_revise_nobdl} is ambiguous and presents some artifacts, and Figure \ref{fig:5_revise_nobdl} does not look like any number at all but rather an arbitrary symbol that fools the classifier into believing it is a 1.

We checked whether an ensemble approach such as the one used by \citet{Schut21}, when applied to REVISE, could improve the counterfactual quality. The idea is to implicitly minimize the epistemic uncertainty of the counterfactual. The new objective function is now given by: 
\begin{align}
   \mathcal{L}_{Revise-e}(z') = &\frac{1}{M} \sum_{m=1}^M L(C_m(\mathcal{G}_{\psi}(z')),y') \nonumber\\
   &+ \lambda \, d(X,\mathcal{G}_{\psi}(z')),
\label{eq:ce_optimization_revise_ensemble}
\end{align}
where $(C_m)_{m=1}^M$ denotes the ensemble of classifiers. This modification will be referred to as REVISE-ENSEMBLE, or REVISE-E.

Figures \ref{fig:6_revise} and \ref{fig:5_revise} show the qualitative results of this modification, with an ensemble of $M=50$ classifiers. The improvements, if any, are marginal in the realism of these counterfactuals.

\section{THE LATENT SPACE AS A BASIS FOR CLASSIFICATION MODELS}
\label{section:encoder_classifier}

In this section, we propose to show how using a classifier directly trained onto the latent space of a VAE leads to higher quality explanations, and we give some insights on why this is the case.

Similarly to REVISE, this new method relies on a latent space from a VAE for producing explanations.
However, instead of using a classifier trained on the image space $\mathcal{X}$, it uses a model trained directly onto the latent space $\mathcal{Z}$. Equation \ref{eq:ce_optimization_encoder_classifier} describes the new objective function
\begin{align}
   \mathcal{L}_{latent}(z') =  L(C(z'),y') + \lambda \, d(z,z'),
\label{eq:ce_optimization_encoder_classifier}
\end{align}
with $z\sim q_\theta(z|X)$ being sampled from the encoder.

Notice that the distance is now computed on the latent space rather than on the image space. While we have not found much difference empirically on MNIST dataset, this ``semantic'' distance (rather than the pixel-wise distance) yields better results on the CelebA dataset \citep{Subramanian15}. See Section \ref{sec:clarity} for more details on this aspect.

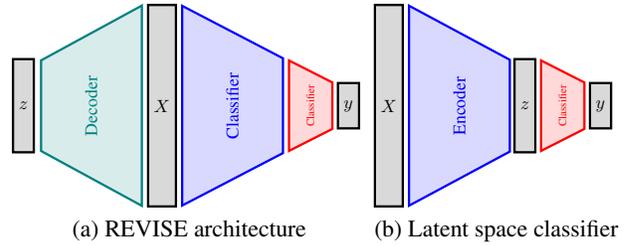
\begin{figure}[htbp]
\centering
\subfloat[REVISE architecture\label{fig:revise_architecture}]{%
\begin{tikzpicture}[scale=0.67, every node/.style={scale=0.67}]
    \node[draw, trapezium, rotate=90, minimum width = 4cm, minimum height = 2cm, trapezium angle = 29, trapezium stretches = true, shape border rotate=360, name=decoder, color=teal, line width=0.3mm, fill=teal!15] at (0,1.5)  {Decoder};
    \node[draw, trapezium, rotate=90, minimum width = 4cm, minimum height = 2cm, trapezium angle = 29, trapezium stretches = true, shape border rotate=180, name=classifier1, color=blue, line width=0.3mm, fill=blue!15] at (2.8,1.5)  {Classifier};
    \node[draw, trapezium, rotate=90, minimum width = 4cm, minimum height = 2cm, trapezium angle = 30, trapezium stretches = true, shape border rotate=180, name=classifier2, color=red, line width=0.3mm, ,scale=0.43, fill=red!15] at (4.35,1.5) {\Large Classifier};
    
    \node (rect) at (-1.35,1.5) [draw,fill=black!15, line width=0.3mm, minimum width=0.2cm,minimum height=1.85cm] {$z$};
    \node (rect) at (1.4,1.5) [draw,fill=black!15, line width=0.3mm, minimum width=0.2cm,minimum height=4cm] {$X$};
    \node (rect) at (5.10,1.5) [draw,fill=black!15, line width=0.3mm, minimum width=0.2cm,minimum height=0.9cm] {$y$};
\end{tikzpicture}
} \hfil
\subfloat[Latent space classifier \label{fig:clarity_architecture}]{%
\begin{tikzpicture}[scale=0.67, every node/.style={scale=0.67}]
    \node[draw, trapezium, rotate=90, minimum width = 4cm, minimum height = 2cm, trapezium angle = 30, trapezium stretches = true, shape border rotate=180, name=encoder, color=blue, line width=0.3mm, fill=blue!15] at (0,1.5)  {Encoder};
    \node[draw, trapezium, rotate=90, minimum width = 4cm, minimum height = 2cm, trapezium angle = 30, trapezium stretches = true, shape border rotate=180, name=classifier, color=red, line width=0.3mm, ,scale=0.43, fill=red!15] at (2.05,1.5) {\Large Classifier};
    \node (rect) at (-1.4,1.5) [draw,fill=black!15, line width=0.3mm, minimum width=0.2cm,minimum height=4cm] {$X$};
    \node (rect) at (1.3,1.5) [draw,fill=black!15, line width=0.3mm, minimum width=0.2cm,minimum height=1.85cm] {$z$};
    \node (rect) at (2.8,1.5) [draw,fill=black!15, line width=0.3mm, minimum width=0.2cm,minimum height=0.9cm] {$y$};
\end{tikzpicture}
} \hfil
\caption{Difference of architecture between REVISE and using a latent space classifier. Blocks of same color share the same architecture.}
\label{fig:architecture}
\end{figure}
We show in Figure \ref{fig:architecture} the architectural difference between REVISE and our approach. 
REVISE uses a classifier in which both blue and red parts are trained for the classification task, whereas we only train the red part after the encoder of the VAE. In the end, the architecture of the classifier for REVISE and the ``Encoder + Classifier'' for our method is the same: they share the same amount of parameters, for fairness reasons. On the one hand, our approach directly computes the gradient $\frac{\partial \mathcal{L}}{\partial z}$ w.r.t. $z$, with $\mathcal{L}$ being the objective function to optimize. On the other hand, REVISE has to go all the way through the decoder: $\frac{\partial \mathcal{L}}{\partial z} = \frac{\partial \mathcal{L}}{\partial X} \frac{\partial X}{\partial z}$. Therefore, we can achieve a faster computation of the gradient compared to REVISE, up to an order of magnitude faster in practice.

\begin{figure}[htbp]
\centering
\subfloat[$6 \rightarrow 3$\label{fig:2_encoder_classifier}]{%
\includegraphics[scale=0.27, trim=40 13 90 40, clip]{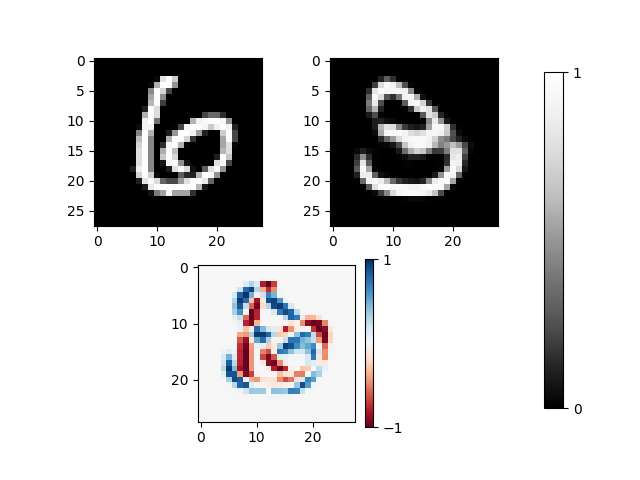}
}\hfil
\subfloat[$5 \rightarrow 1$\label{fig:5_encoder_classifier}]{%
\includegraphics[scale=0.27, trim=40 13 40 40, clip]{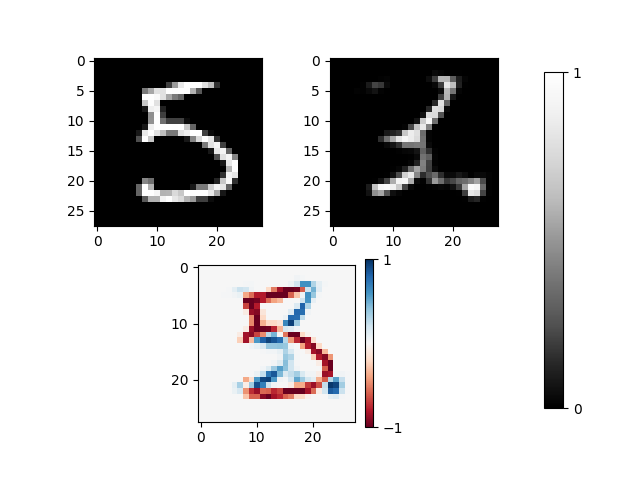}
}\hfil\\
\vspace{.1in}
\caption{Counterfactual explanations using a latent space classifier}
\label{fig:mnist_encoder_classifier}
\end{figure}

Figure \ref{fig:mnist_encoder_classifier} shows the qualitative results of our approach on MNIST. 
Just like the image space classifiers, we use adversarial training. The explanations are generated using $\lambda=0.001$, and stop at a target probability of $0.99$. As we can see, the counterfactual explanation $6 \rightarrow 3$ is very realistic as it presents no artifacts nor ambiguity. The explanation even preserves features such as handwriting since the thickness of the stroke is the same in both the original image and the counterfactual explanation.
However, the transformation $5 \rightarrow 1$ is not quite  satisfactory, as it still presents  artifacts and ambiguity.

\begin{figure}[htbp]
\centering
\subfloat[$6 \rightarrow 3$\label{fig:63_clarity_revise}]{%
\includegraphics[scale=0.27, trim=20 40 40 80, clip]{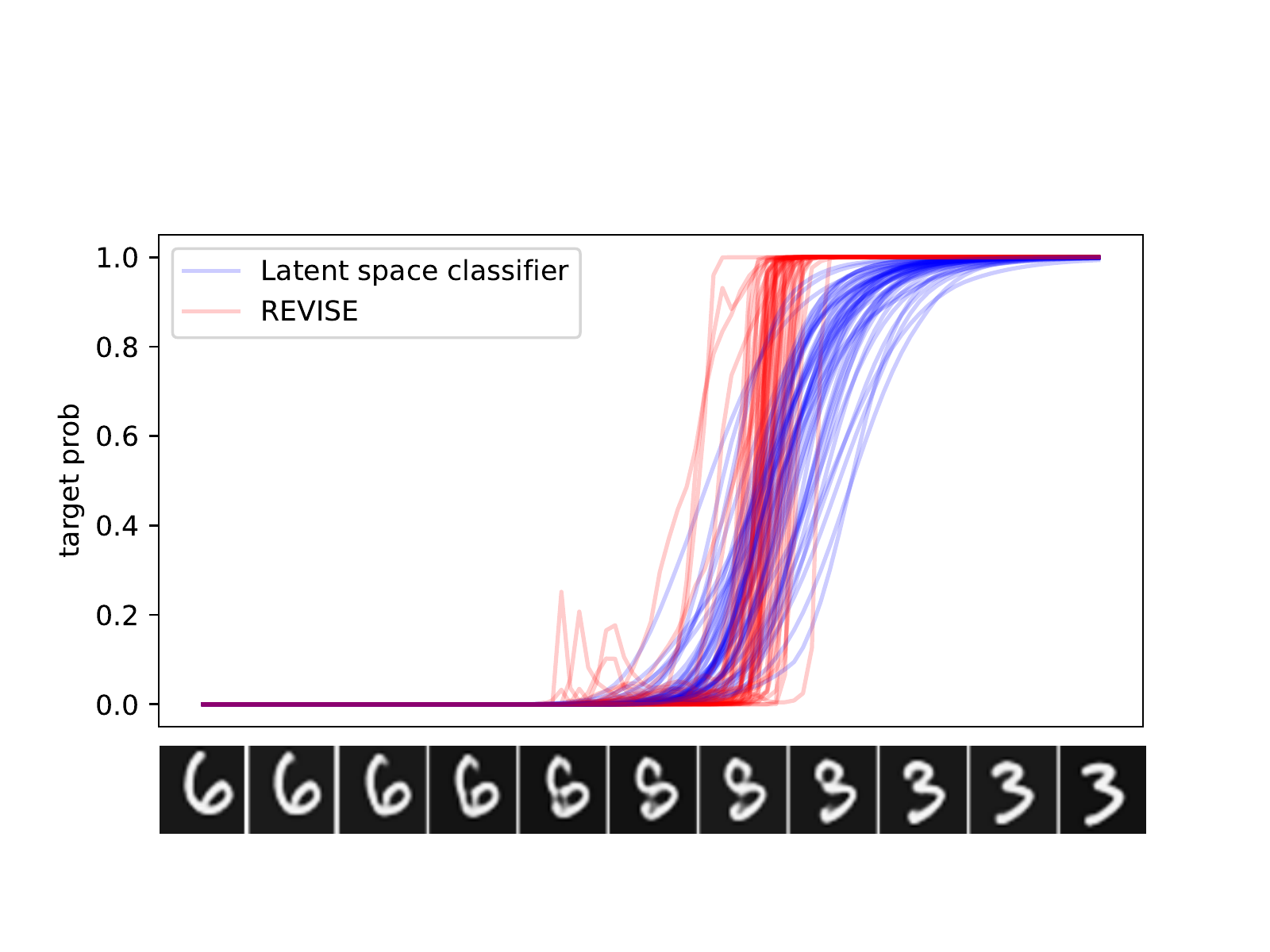}
}\hfil
\subfloat[$5 \rightarrow 1$\label{fig:28_clarity_revise}]{%
\includegraphics[scale=0.27, trim=20 40 40 80,clip]{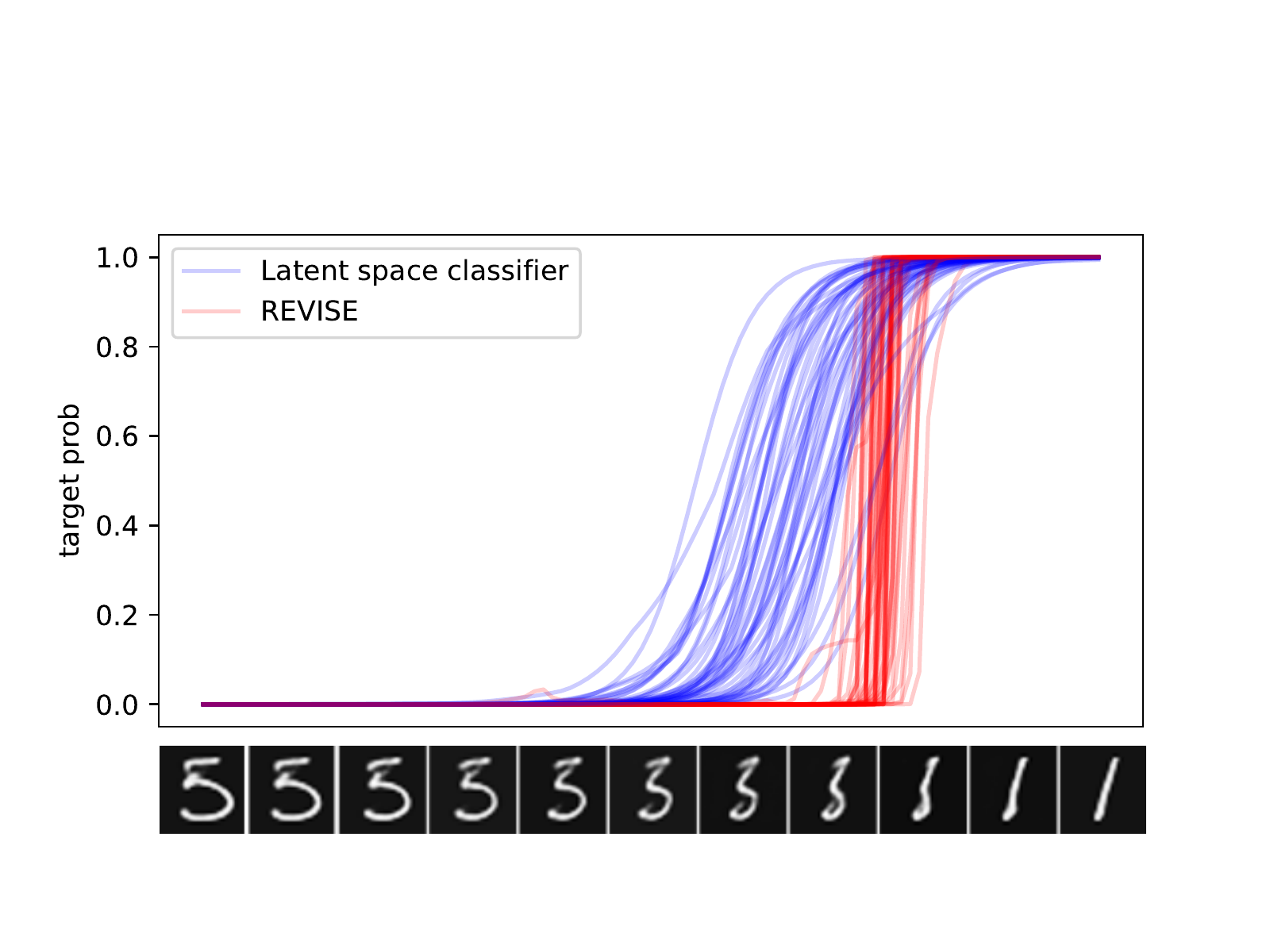}
}\hfil
\vspace{.1in}
\caption{Probability of the target class with respect to the interpolation in the latent space. In red: REVISE. In blue: latent space classifier.}
\label{fig:interpolation_clarity_revise}
\end{figure}

Given these results, one might ask how training a model directly on a VAE latent space can achieve more realistic explanations. In Figure \ref{fig:interpolation_clarity_revise}, we interpolate two images of two different classes into the latent space given the equation $z(t) = (1-t)\, z_1+t \, z_2, \, t\in[0,1]$. Both $z_1$ and $z_2$ are latent representations from the encoder of the VAE of the two images $X_1$ and $X_2$, with distinct respective classes $y_1$ and $y_2$. We also plot the target probability $t \mapsto P(Y=y_2 \,|\, z(t))$ for both the image and the latent space classifiers. For the image space classifier, we first decode $X(t)=\mathcal{G}_{\psi}(z(t))$ and then compute the predictive probability of $X(t)$. For each type of model, we trained 50 of them and plotted their curves in order to make sure the behavior observed is independent of the epistemic variance.

\begin{figure}[htbp]
\centering
\captionsetup[subfigure]{justification=centering}
\subfloat[][\centering Classifier trained and viewed from image space.\label{fig:image_space_interpolation}]{%
\includegraphics[scale=0.32]{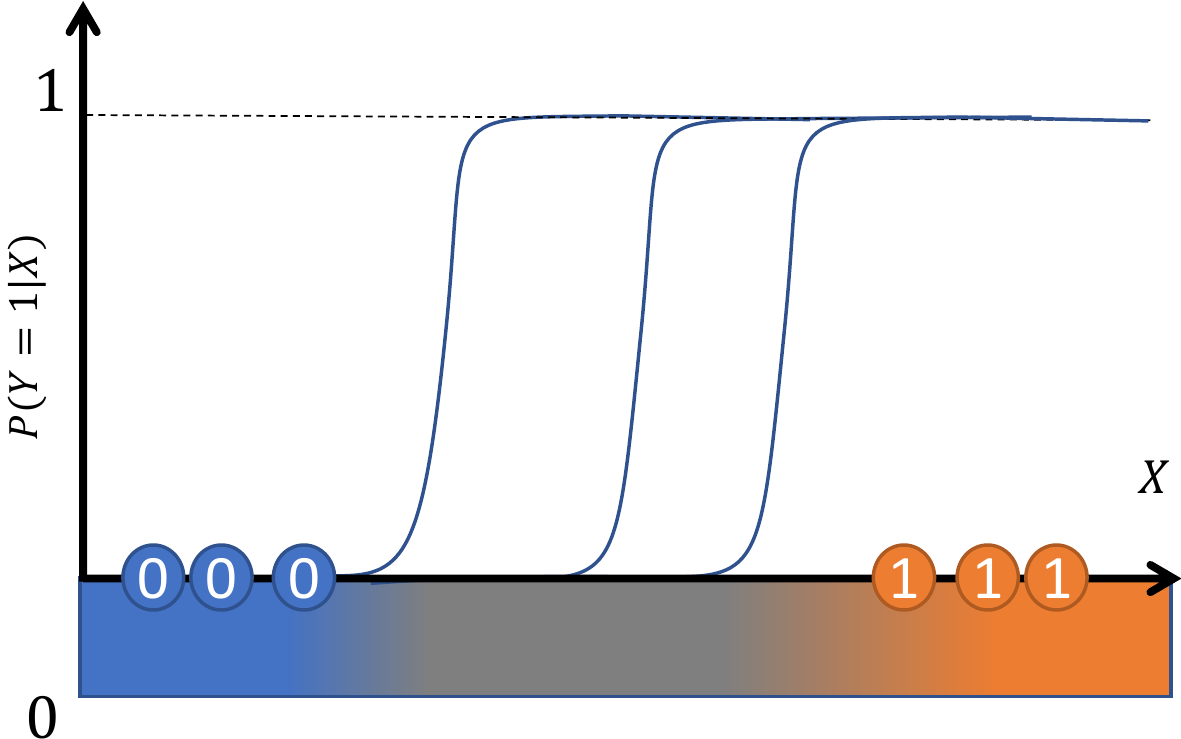}
}\hfil
\subfloat[][\centering Classifier trained in image space, viewed from latent space.\label{fig:latent_space_interpolation}]{%
\includegraphics[scale=0.32]{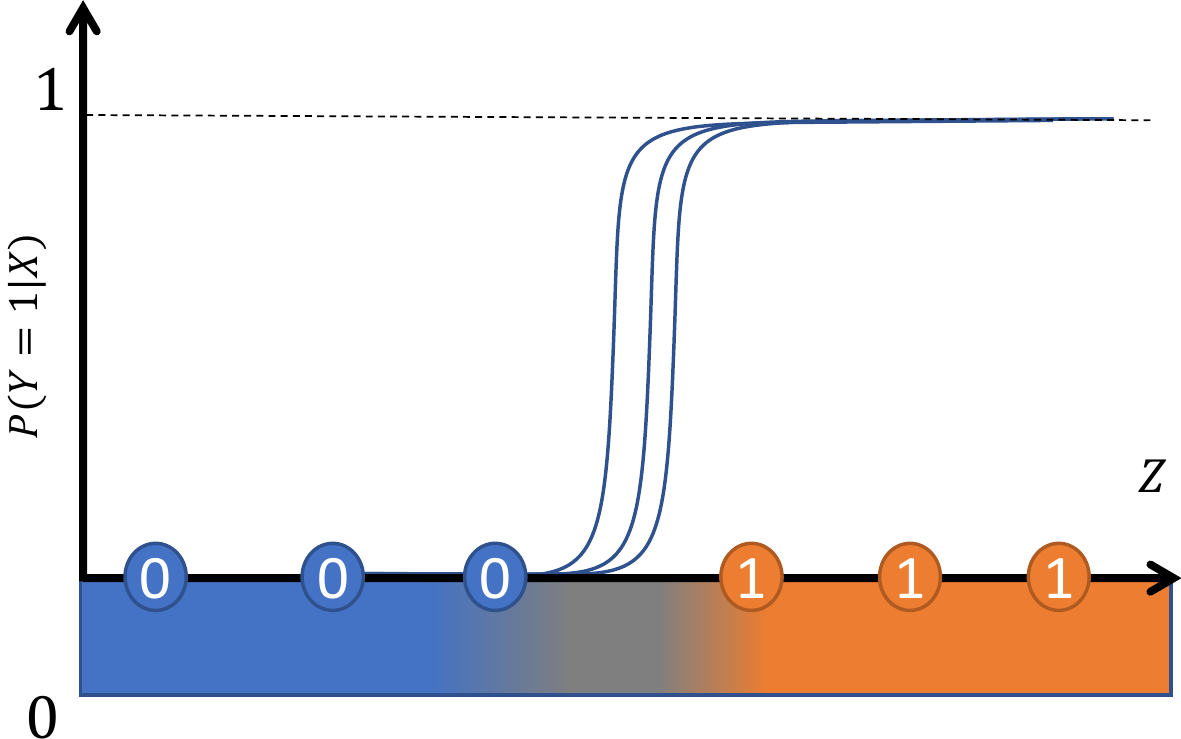}
}\hfil\\
\subfloat[][\centering Classifier trained and view from latent space.\label{fig:clarity_interpolation}]{%
\includegraphics[scale=0.32]{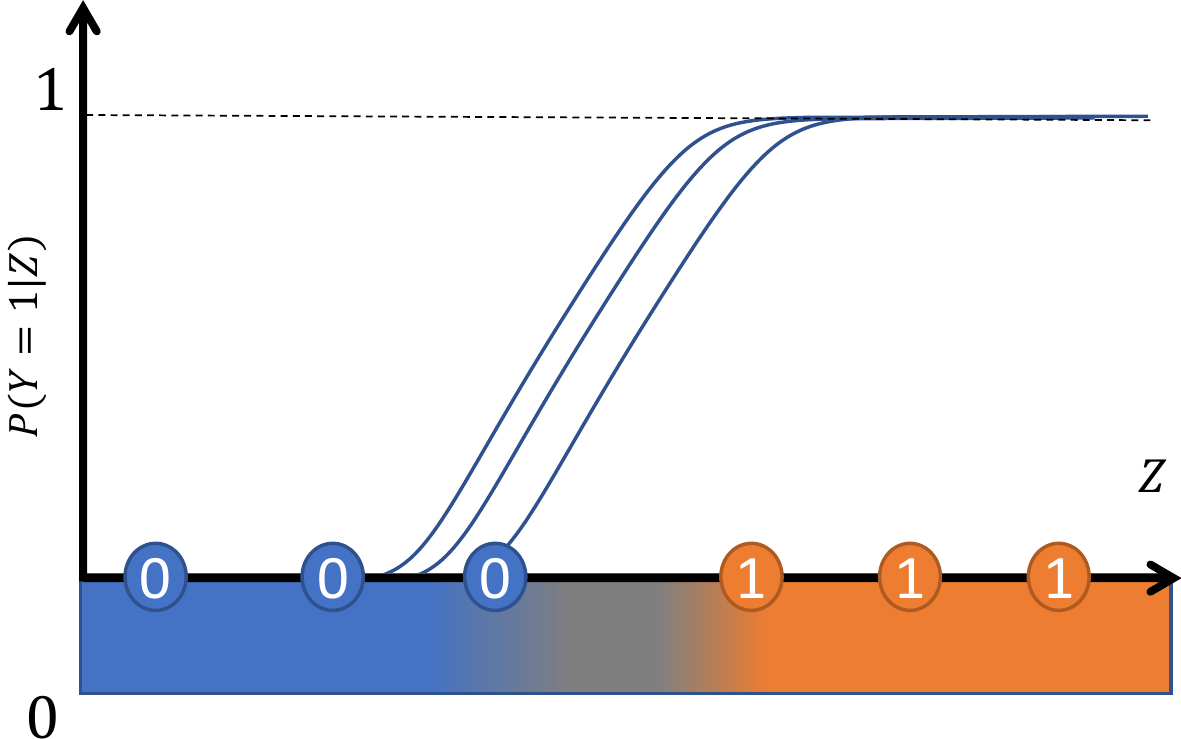}
}\hfil\\
\vspace{.1in}
\caption{Illustration of the probability of the target class $Y=1$ for three different classifiers (representing the epistemic uncertainty), in multiple settings.
The 1D x-axis represents either the high dimensional image or latent space. Examples of classes 0 and 1 are represented as circles on the x-axis. The grey color in the x-axis gradient represents the area between the two classes with out-of-distribution, unrealistic images, which is compressed in the latent space.}
\label{fig:interpolation_figure}
\end{figure}

Notice how steep the curves for the image classifiers are when interpolating the latent space, in contrast to the classifiers directly trained in it.
To understand the reason for this, let us recall that the density induced by the training data is extremely sparse when expressed in the image space: each class forms an ``isolated island'' of density.
To go from one class to another, it is therefore necessary to cross spaces of very low density, corresponding to out-of-distribution non-realistic images.
A classifier trained directly in the image space will therefore have the possibility to arbitrarily place separating boundaries between classes within this space, as illustrated on Fig.~\ref{fig:image_space_interpolation}. 
Moreover, the cross-entropy will encourage transitions to be as steep as possible, unless there is a strong regularization on weights.
Consequently, outside of these transition zones, the probability of the target class will be particularly stable, often equal to $0+\varepsilon$ or $1-\varepsilon$. This is exactly the ``plateau effect'' observable on the left and right-hand sides of the curves in Figure \ref{fig:interpolation_clarity_revise}.

\begin{figure}[htbp]
\centering
\subfloat[Image space interpolation\label{fig:interpolation_image}]{%
\includegraphics[scale=0.27, trim=20 40 40 80, clip]{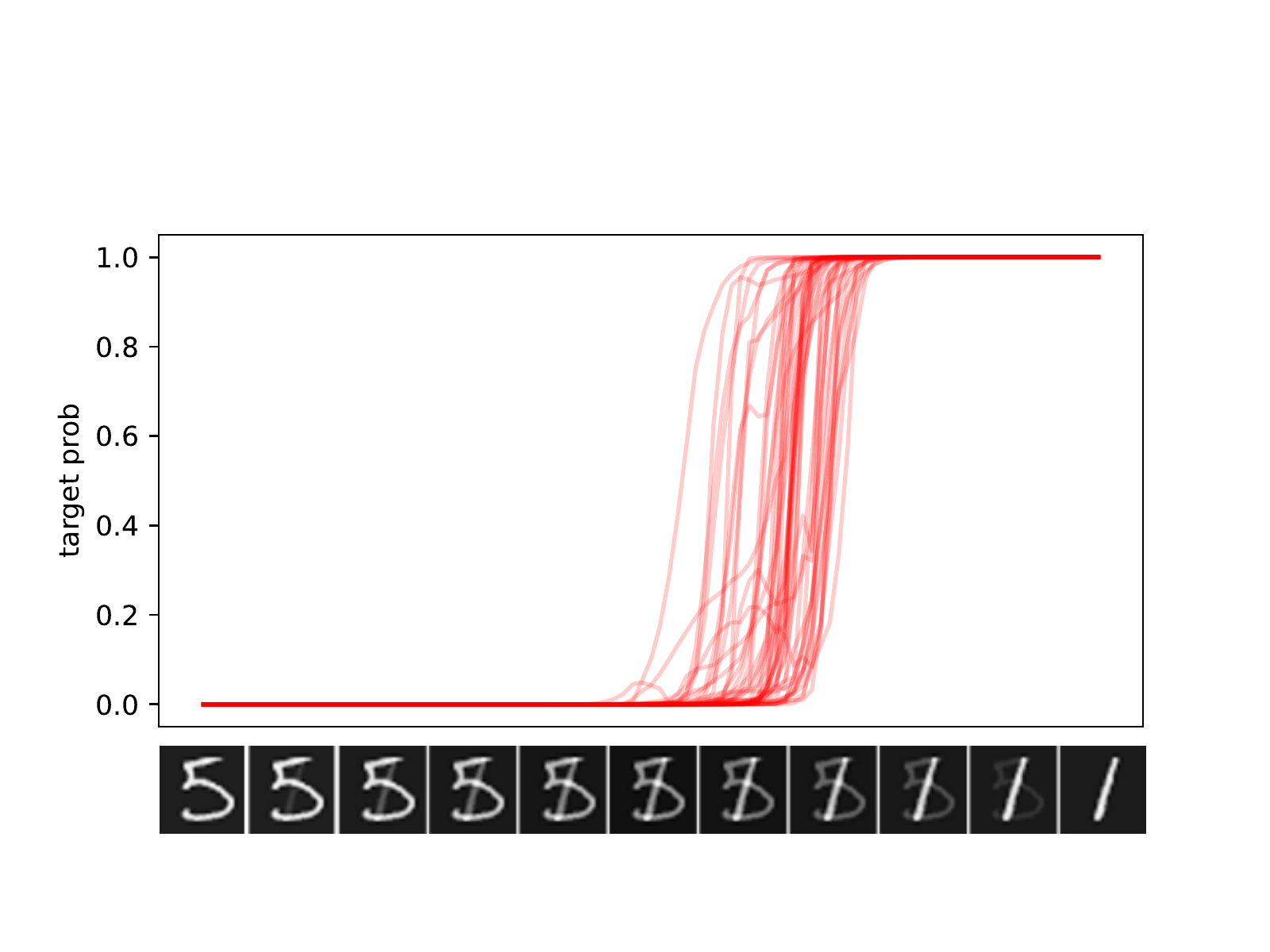}
}\hfil
\subfloat[Latent space interpolation\label{fig:interpolation latent}]{%
\includegraphics[scale=0.27, trim=20 40 40 80,clip]{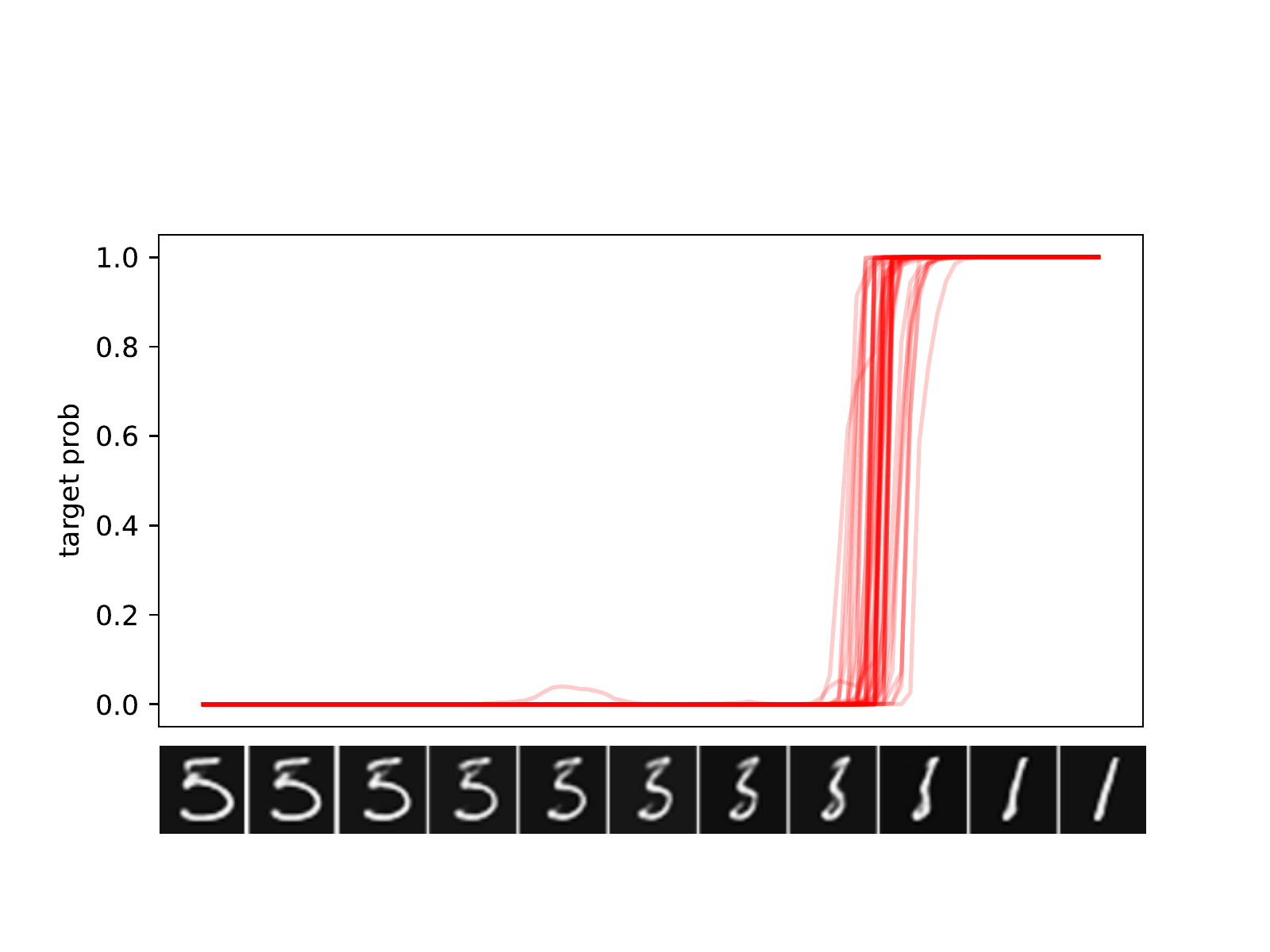}
}\hfil
\vspace{.1in}
\caption{Probabilities of the target class from the ensemble of image space classifiers, with respect to the interpolation in the image and latent space respectively, from 5 to 1.}
\label{fig:interpolation_comparison}
\end{figure}

How do these facts translate into the VAE's latent space? During training, the VAE compresses the information contained in the density underlying the training data in such a way that the high density subspaces occupy a preponderant part of the latent space and, on the contrary, that the low density areas are mapped to very small volumes.
The distribution of each class of images being represented by numerous examples (e.g., many characteristic images 
in MNIST) is thus mapped to significant volumes of the latent space, whereas the inter-class transition zones, corresponding to unrealistic images not observed during training, are reduced to very thin layers of the latent space, as illustrated on Fig.~\ref{fig:latent_space_interpolation}.
Experimental results shown on Figure~\ref{fig:interpolation_comparison} confirm this phenomenon: if we consider the interpolated segments of images either in the image or in the latent space, we see that the probability transitions of classifiers trained in the image space clearly appear more tightened in the latent space than in the image space, reflecting the compression effect on low density areas.
Since the probability transition areas are very small when viewed from the latent space, the smoothing of the probabilities by a ensemble of classifiers is very localized and thus rendered inefficient, which explains why REVISE-ENSEMBLE does not add much compared to REVISE.

As a consequence, the VAE distorts geometry in the latent space in such a way that unrealistic transition areas can appear very close to realistic image plateaux, attracting the gradient in a wrong direction, all the more easily as probability transitions are steep.
Once the transition is underway, the probability of the target class reaches the extreme value of $1$ very quickly so that the gradient descent stops, without the corresponding counterfactual being realistic.

On the other hand, by training the classifier directly in the latent space, the model is forced to be constrained by the lower dimension, which leads to smoother separations less prone to overfitting, as illustrated in Fig.~\ref{fig:clarity_interpolation}. We can clearly see in Fig. \ref{fig:interpolation_clarity_revise} how the model trained in latent space isn't confident as long as the example is ambiguous. This can explain why latent-space based classifiers are more interpretable, as the gradients lead more naturally to examples closer to the target class and not outliers. 

\begin{figure}[htbp]
\centering
\subfloat[REVISE-E\label{fig:convergence_revise}]{%
\includegraphics[scale=0.245, trim=0 0 0 0, clip]{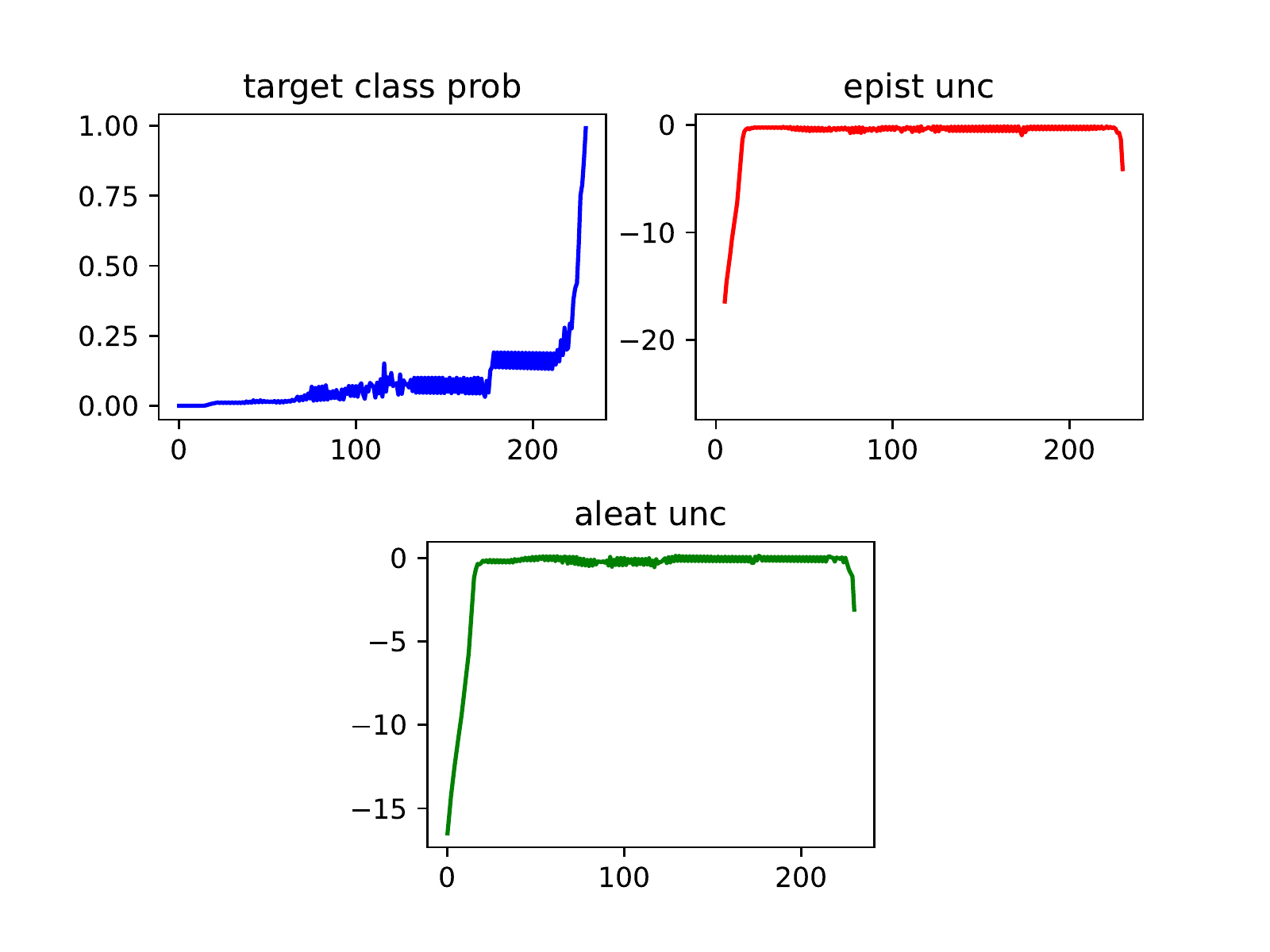}
}\hfil
\subfloat[Clarity\label{fig:convergence_clarity}]{%
\includegraphics[scale=0.245, trim=0 0 0 0,clip]{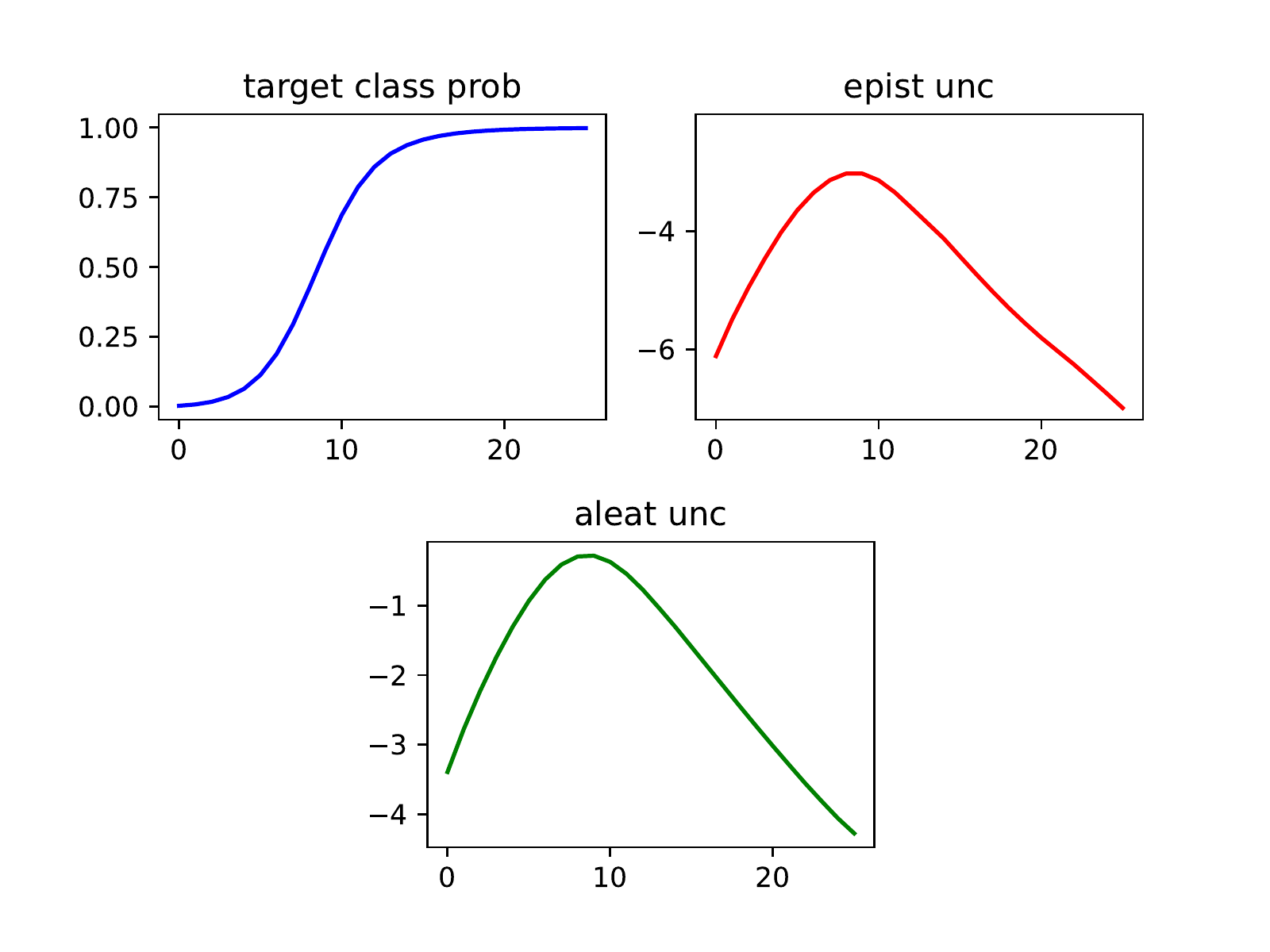}
}\hfil\\
\subfloat[3 to 8 (REVISE-E)\label{fig:3_revise}]{%
\includegraphics[scale=0.28, trim=40 13 90 40, clip]{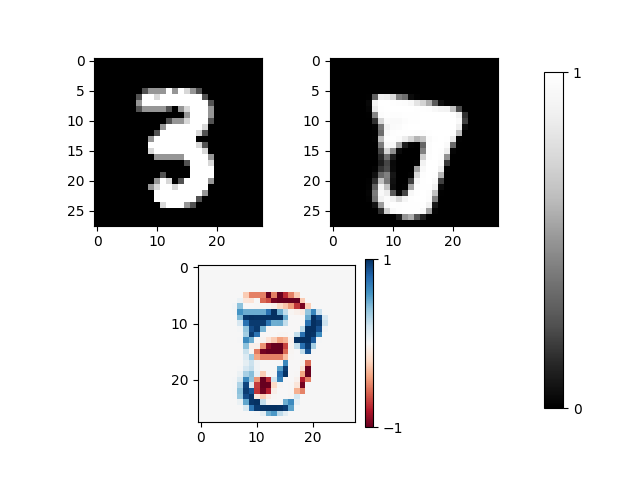}
}\hfil
\subfloat[3 to 8 (Clarity)\label{fig:3_clarity}]{%
\includegraphics[scale=0.28, trim=40 13 90 40, clip]{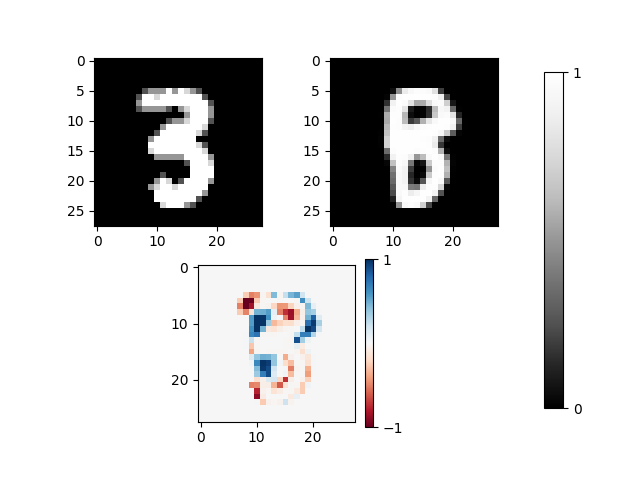}
}\hfil\\
\vspace{.1in}
\caption{Probability curves of the target class (blue), logarithm of epistemic uncertainty (red) and aleatoric uncertainty (green) along the steps of the counterfactual explanation generation, for both Clarity and REVISE-E. The explanation generated is $3 \rightarrow 8$.}
\label{fig:convergence_clarity_revise}
\end{figure}

We can also notice in Fig. \ref{fig:63_clarity_revise} how, in the case of REVISE, the probability curve is not monotonous, with multiple local maxima. These instabilities again illustrate how the image space classifier is not well calibrated when observed from the latent space. They also entail the risk for REVISE of being stuck in local maxima when trying to generate an explanation. This has been noticed in a few cases, like in Fig.~\ref{fig:3_revise}  where we ask REVISE-E to come up with a counterfactual explanation from a 3 to an 8.  REVISE-E struggles for more than 200 steps to get out of a local maximum and fails by returning an unrealistic counterfactual. This is contrasted by the results of the latent space trained classifiers, which not only yield smooth curves, but monotonous as well. 
However, we can notice in Figure~\ref{fig:interpolation_clarity_revise} the variance between the different classifiers in the interpolated probability curves. We will explain in the next section how leveraging an ensemble of models to minimize the epistemic uncertainty (in addition to the latent space classifiers) can lead to better explanations.

\section{CLARITY: AN EXPLAINABLE BAYESIAN LATENT SPACE CLASSIFIER}
\label{sec:clarity}

We will now present \textit{Clarity}, a type of classifier that subsumes every component we have seen so far to produce quality visual counterfactual explanations. The basis behind \textit{Clarity} is to use a Bayesian classifier trained on a latent space of a VAE. In Section \ref{section:encoder_classifier}, we have seen how using a latent space classifier helps to produce classifiers that are less overconfident on ambiguous examples, due to the fact that the latent space, being lower-dimensional and continuous, removes sparsity in the input space. However, in Figure \ref{fig:interpolation_clarity_revise}, we can see that, given an ensemble of latent space classifiers $(C_m)_{m=1}^M$, there is a noticeable variance on how each of the classifiers predict its target probability. In order to minimize the epistemic uncertainty during the process of generating a counterfactual explanation, we propose to use an ensemble of classifiers just like in \cite{Schut21}.

The new 
objective function is given Equation \ref{eq:ce_optimization_clarity}. We choose as a starting point the mean of the variational posterior, and thus our algorithm is deterministic. 
\begin{align}
 \mathcal{L}_{Clarity}(z') = \frac{1}{M} \sum_{m=1}^M L(C_m(z'),y') + \lambda \, d(z,z').
\label{eq:ce_optimization_clarity}
 \end{align}
Following this new objective, we define our counterfactual generation algorithm in Algorithm \ref{alg:alg_1}.
\begin{algorithm}[H]
\SetAlgoLined
\textbf{Input:} Original image $X$, target class $y'$, target probability $\gamma$, maximum number of iterations $N$, hyperparameter $\lambda$, VAE with variational posterior $q_\theta(z|X)=\mathcal{N}(\mu,\Sigma)$ and decoder $\mathcal{G}_{\psi}(z)$, ensemble of classifiers $(C_m(z))_{m=1}^M$, optimizer \textit{opt}\\
\textbf{Output:} counterfactual image $X'$\\
$\mu, \Sigma \leftarrow q_\theta(z|X)$ \\
$z \leftarrow \mu$, $z' \leftarrow z$, $i \leftarrow 0$ \\
\While{$\frac{1}{M}\sum_{m=1}^M p(y'\,|\,z')\leq\gamma$ and $i \leq N$}{
    $S(z',y') = \nabla_{z'} \left(\frac{1}{M} \sum\limits_{m=1}^M L(C_m(z'),y') + \lambda \, d(z,z') \right)$\\
    $z' \leftarrow opt(z',S(z',y'))$ \\
    $i \leftarrow i+1$
 }
\Return{$X'=\mathcal{G}_{\psi}(z')$}
\caption{\textit{Clarity} counterfactual example generation}
\label{alg:alg_1}
\end{algorithm}

\begin{figure}[htbp]
\centering
\subfloat[$6 \rightarrow 3$\label{fig:2_clarity}]{%
\includegraphics[scale=0.27, trim=40 13 90 40, clip]{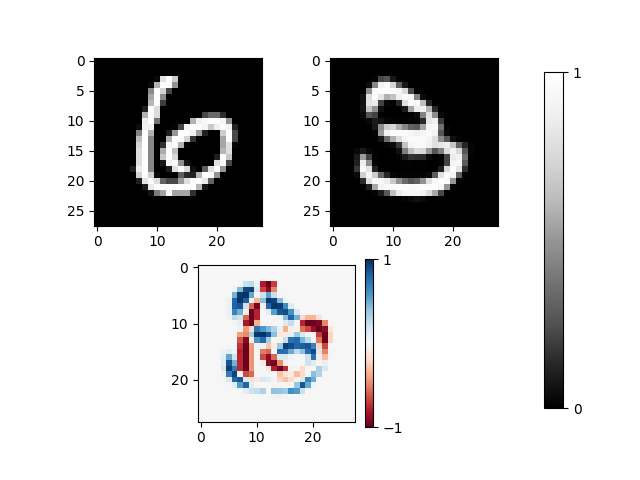}
}\hfil
\subfloat[$5 \rightarrow 1$\label{fig:5_clarity}]{%
\includegraphics[scale=0.27, trim=40 13 40 40, clip]{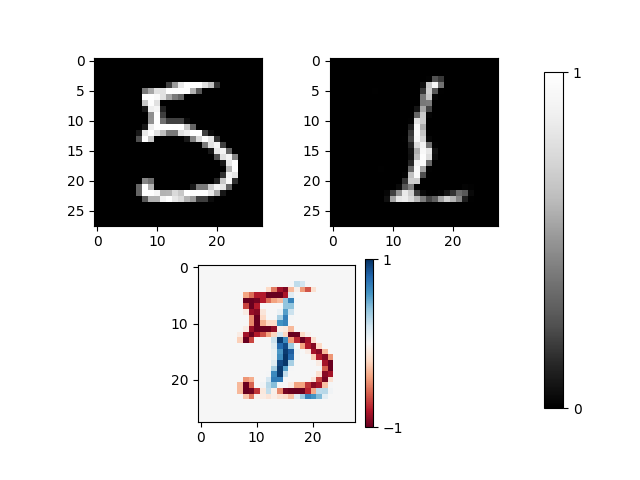}
}\hfil\\
\vspace{.1in}
\caption{Counterfactual explanations using \textit{Clarity}}
\label{fig:mnist_clarity}
\end{figure}

In Figure \ref{fig:mnist_clarity}, we show qualitative results of this new method.
Figure \ref{fig:2_clarity} displays some slight improvements on the counterfactual explanation, as the 3 is more continuous, especially in the middle. However, the most noticeable improvements are seen in Figure \ref{fig:5_clarity}. The counterfactual image is much more realistic when considering it is representing a 1, compared to previous methods.

Returning to the example of Figure~\ref{fig:convergence_clarity_revise}, Clarity produces a more realistic counterfactual than REVISE-E. Clarity is also more consistent than REVISE-E in terms of uncertainty, since with REVISE-E the final uncertainties, both aleatoric and epistemic, are much higher than the uncertainties of the original example, whereas the initial and final uncertainties are around the same value with our method.

\begin{figure}[htbp]
\centering
\subfloat[Gradient descent\label{fig:latent_cnn}]{%
\includegraphics[scale=0.27, trim=20 0 40 20, clip]{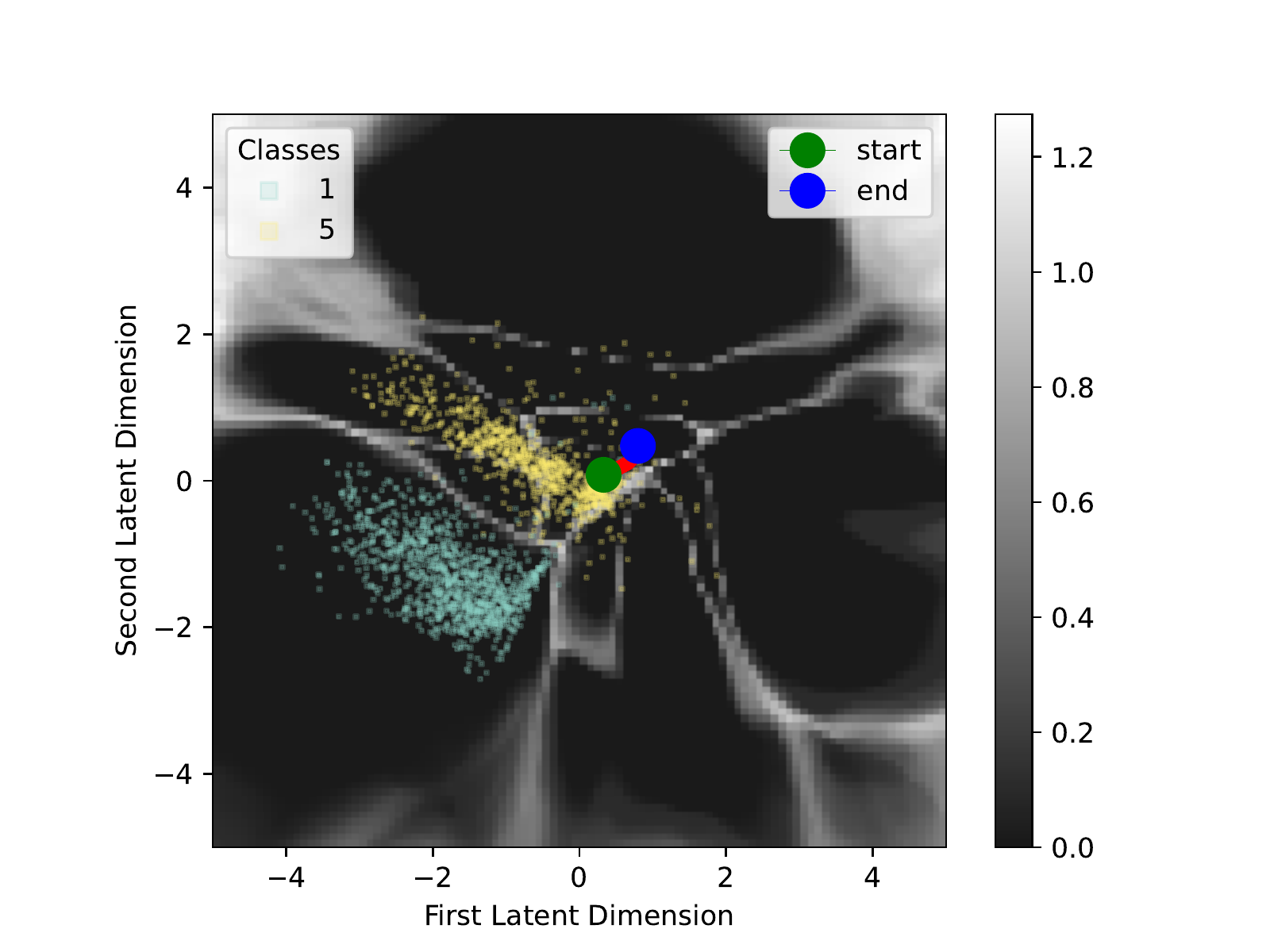}
}\hfil
\subfloat[Schut\label{fig:latent_schut}]{%
\includegraphics[scale=0.27, trim=20 0 40 20, clip]{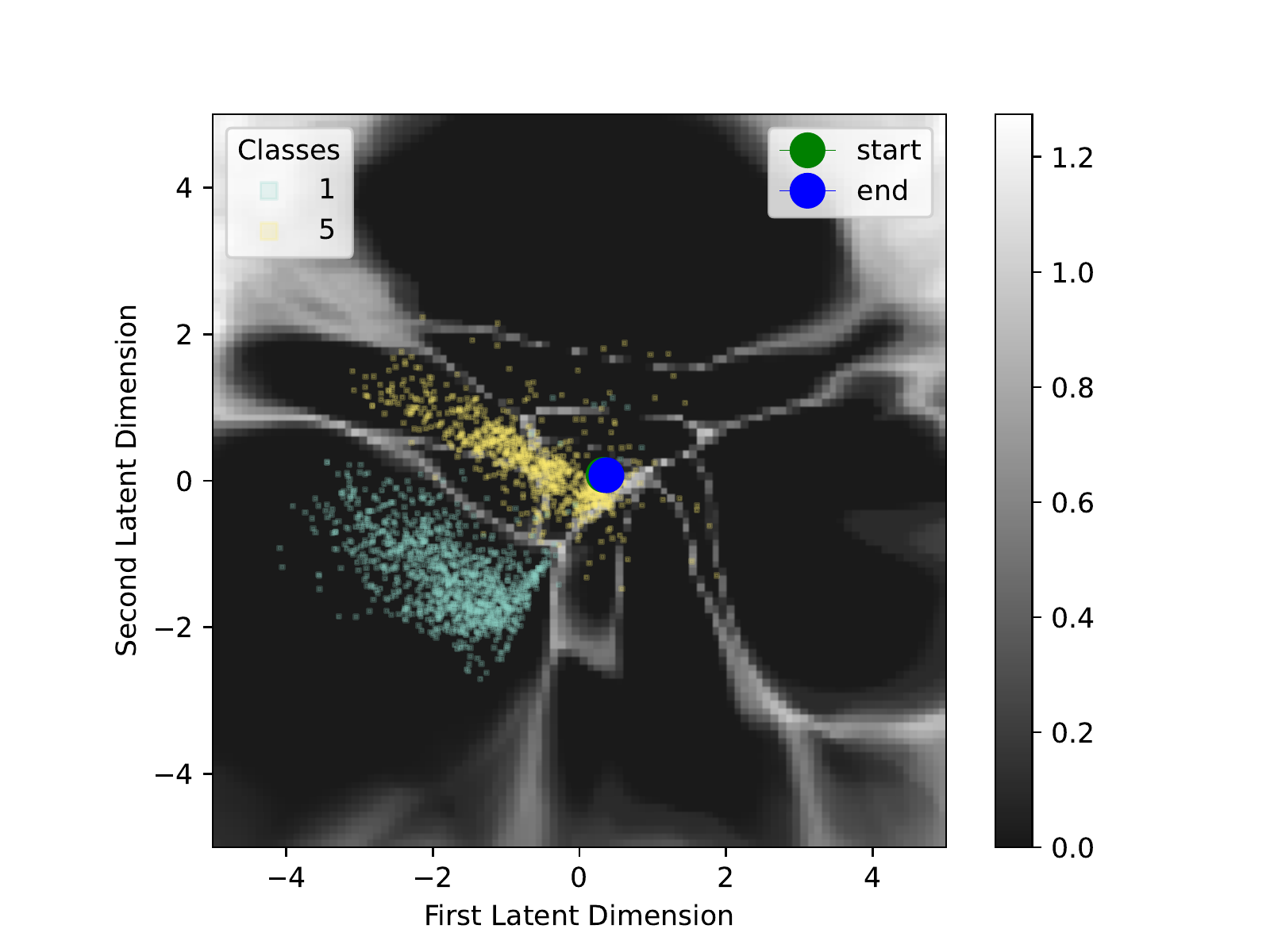}
}\hfil
\subfloat[REVISE-E\label{fig:latent_revise}]{%
\includegraphics[scale=0.27, trim=20 0 40 20, clip]{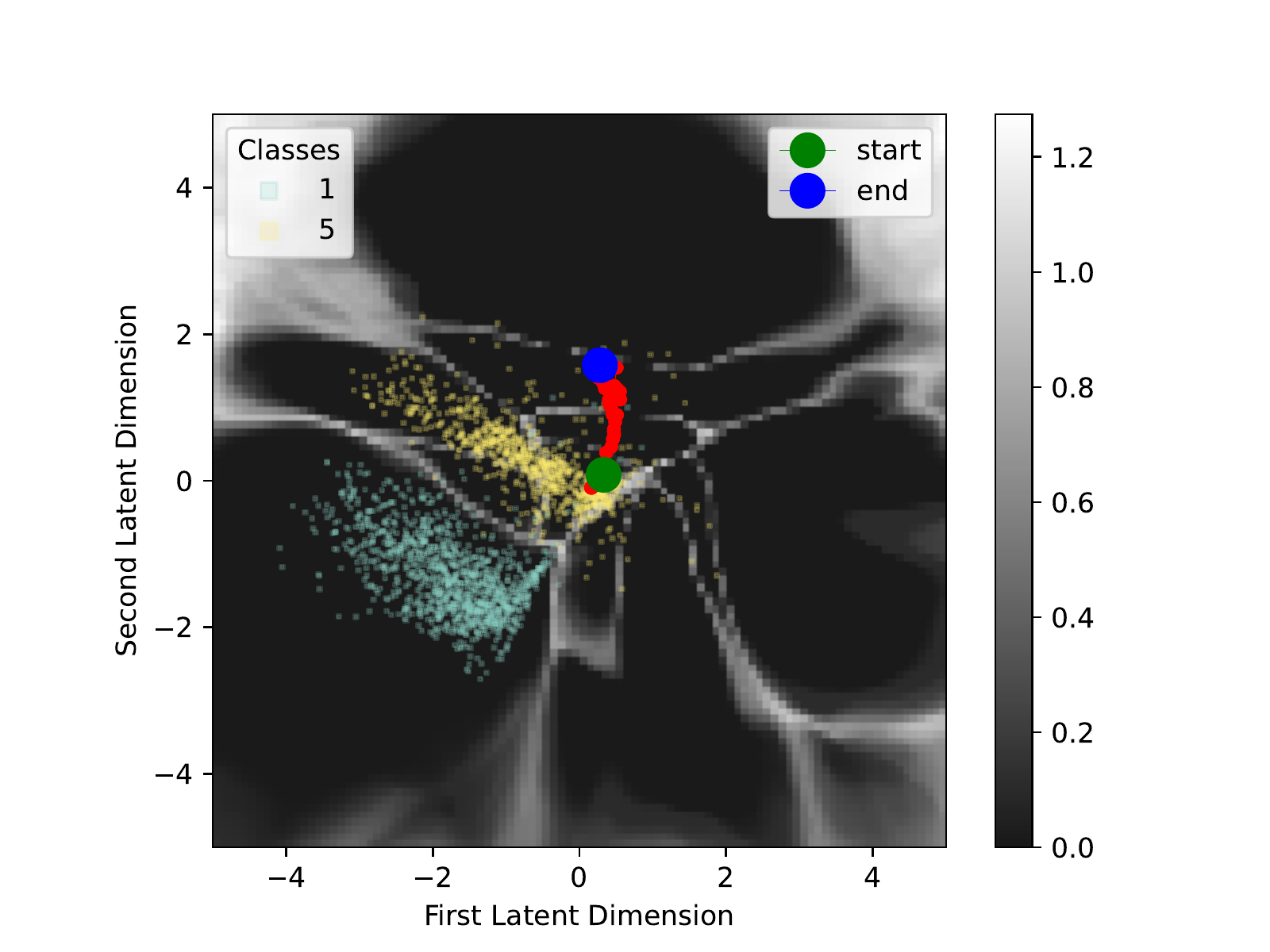}
}\hfil
\subfloat[Clarity\label{fig:latent_clarity}]{%
\includegraphics[scale=0.27, trim=20 0 40 20, clip]{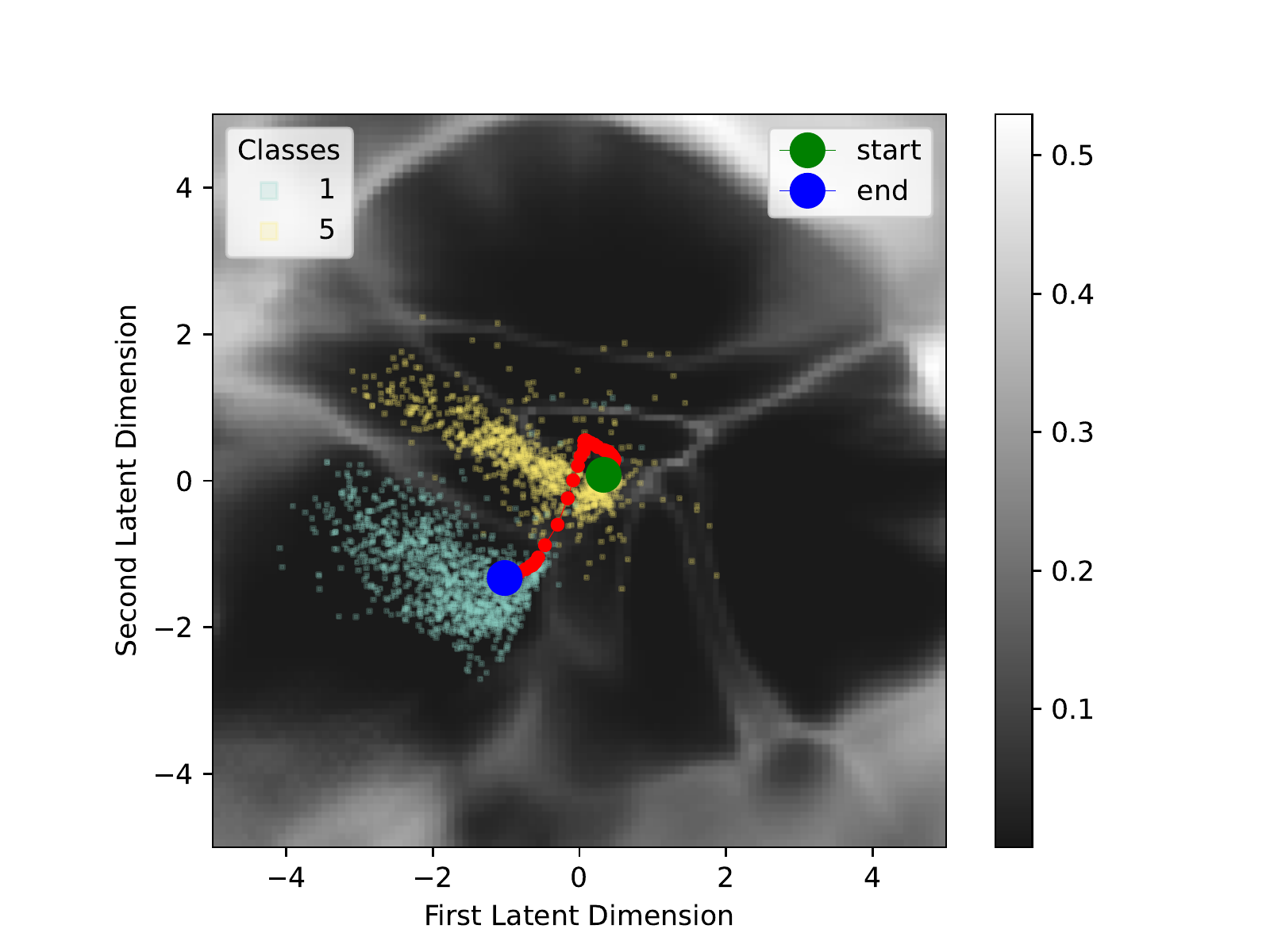}
}\hfil
\vspace{.1in}
\caption{Counterfactual explanation $5 \rightarrow 1$ trajectory in a 2D latent space from a VAE. Green dot is starting point and blue dot is end point. Yellow and cyan points are data points from the test distribution of class 5 and 1 respectively. Grey scale represents epistemic uncertainty. Note that Figure \ref{fig:latent_clarity} has a different uncertainty scale as the classifier is different for Clarity. Zoom for better visualisation.}
\label{fig:latent}
\end{figure}

In addition to those counterfactual images, we show the trajectories of the $5 \rightarrow 1$ counterfactual explanations for each method on a 2D latent space, in Figure \ref{fig:latent}.
This 2D latent space has been generated by another VAE, as described in the work of \cite{Smith18}. The background grey scale represents the epistemic uncertainty of the current classifier (latent space based for \textit{Clarity}, image space based for the others), for each sampled point in the 2D latent space. Firstly, notice how the image space classifier is overconfident in areas that are well beyond the test samples, while the latent space classifier is more conservative with its uncertainty estimates. Secondly, among all counterfactual generation algorithms, the one given by \textit{Clarity} ends up in the middle of the data distribution of the class ``1'' (in cyan), which is what we expected. Image space based algorithms fail to change the image in a meaningful way, as the counterfactual is very close to the original example in the 2D latent space, even though the image has been clearly modified. Finally, REVISE produces an example that is indeed different in the 2D latent space, but that does not end up in the area of the space that interests us and instead gives an example that is far away from the class ``1''.

These empirical results show how the combined techniques given by \textit{Clarity} help to compute a gradient that is more meaningful since the direction of the gradient descent algorithm leads to a realistic example that lies in the data distribution. Training a classifier on a latent space forces the model to learn class separations that are well-fitted in terms of semantic and high-level features. Furthermore, the use of an ensemble of models helps to reduce the variance in the gradient and to follow a direction that ends up minimizing both epistemic and aleatoric uncertainties.

\paragraph{Additional experiments}

 In Table \ref{table:mnist} we show additional empirical results on the MNIST dataset. Each row represents a particular example, and each column a different counterfactual explanation algorithm, with the exception of the first column being the original image. The VAE encoder uses 3 convolutional layers with Batch Normalization \citep{Ioffe15} with PReLU activation \citep{He15}. Then, the latent space is of size 16 and the decoder mirrors the encoder with deconvolution layers. For \textit{Clarity}, the latent space classifier is composed of an input layer of size $16$, two dense layers of size $16$ with ReLU activation, and the final output layer. For all other methods, as explained above, the classifier used has the architecture of the encoder plus the latent space classifier, but now all layers are trained for the classification task. 
 The image space classifiers reach on average an accuracy of around 99\% on the test dataset, while our latent space classifier averages its test accuracy at 96\%.
\begin{table}[htbp]
\caption{Counterfactual explanations comparison on the MNIST dataset. From top to bottom, the explanations are: $3 \rightarrow 8, 0 \rightarrow 9, 4 \rightarrow 7, 8 \rightarrow 0, 2 \rightarrow 1, 9 \rightarrow 5$.}
\begin{center}
  \begin{tabularx}{\linewidth}{*{5}{Xc}}
      \hline Original & FGSM & Schut  & REVISE-E & Clarity \\
      \hline \\
      \includegraphics[scale=0.12]{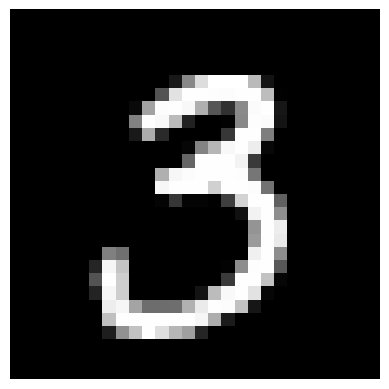} &\includegraphics[scale=0.12]{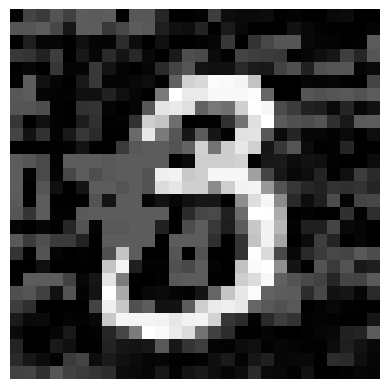} &\includegraphics[scale=0.12]{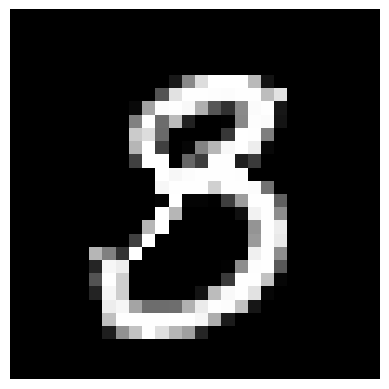} &\includegraphics[scale=0.12]{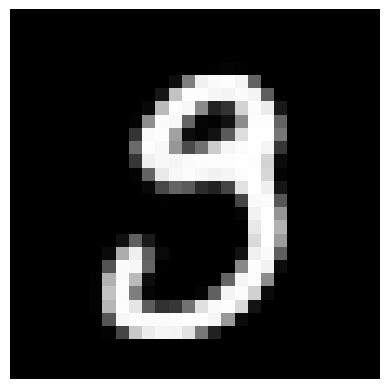}
      &\includegraphics[scale=0.12]{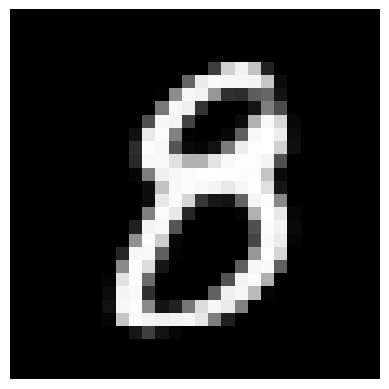}\\
      \includegraphics[scale=0.12]{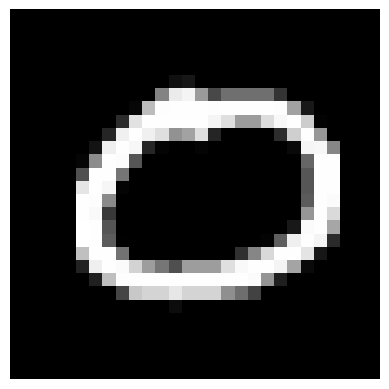} &\includegraphics[scale=0.12]{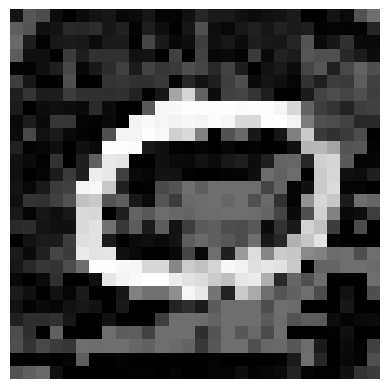} &\includegraphics[scale=0.12]{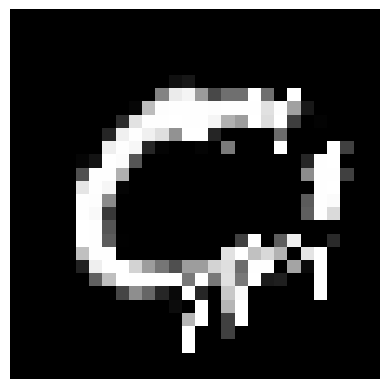} &\includegraphics[scale=0.12]{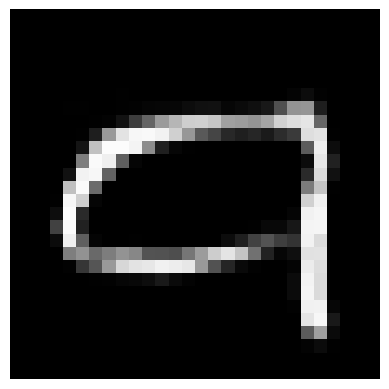}
      &\includegraphics[scale=0.12]{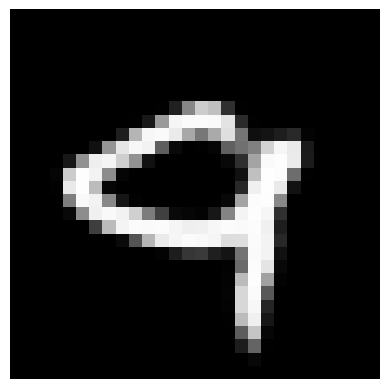}\\
      \includegraphics[scale=0.12]{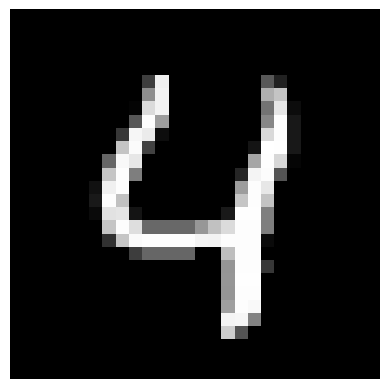} &\includegraphics[scale=0.12]{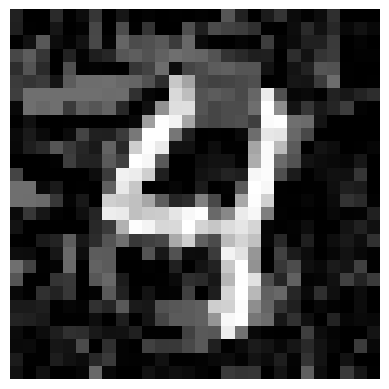} &\includegraphics[scale=0.12]{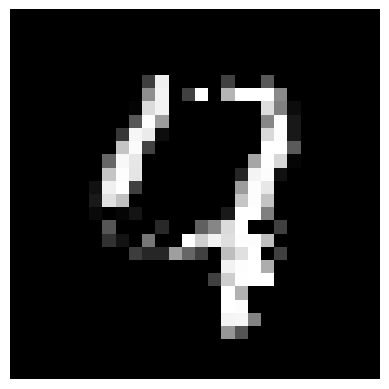} &\includegraphics[scale=0.12]{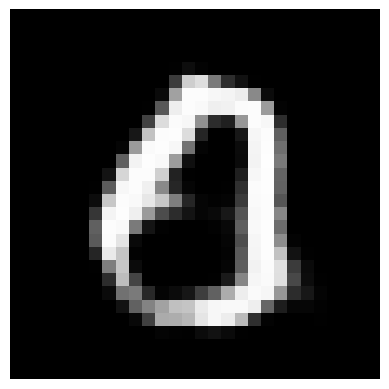}
      &\includegraphics[scale=0.12]{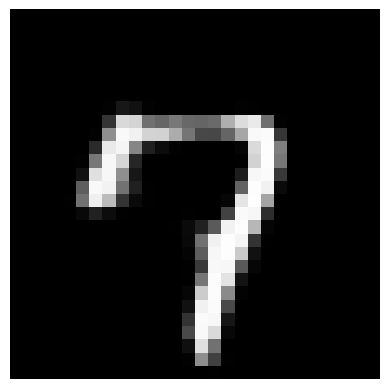}\\
      \includegraphics[scale=0.12]{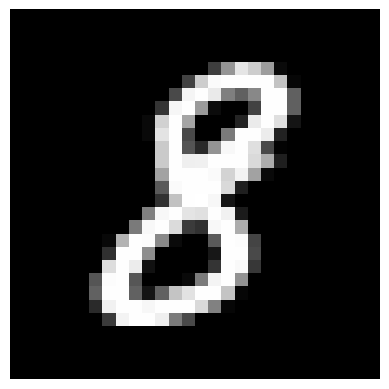} &\includegraphics[scale=0.12]{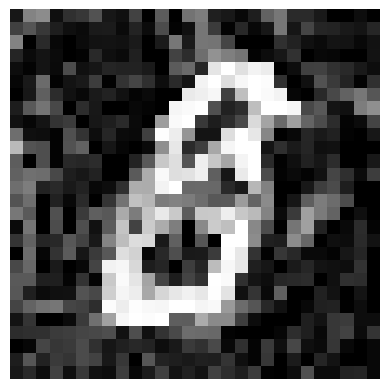} &\includegraphics[scale=0.12]{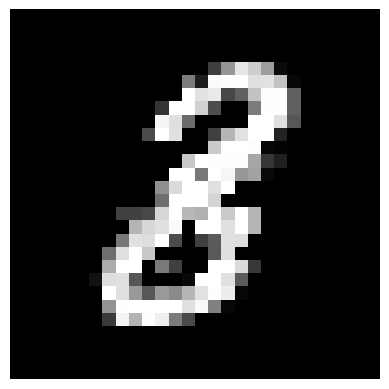} &\includegraphics[scale=0.12]{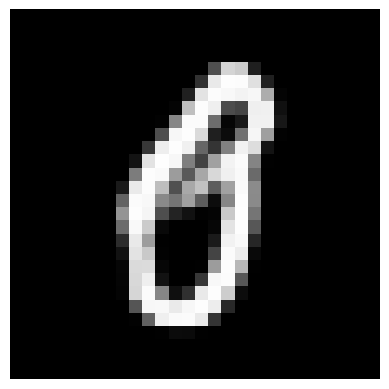}
      &\includegraphics[scale=0.12]{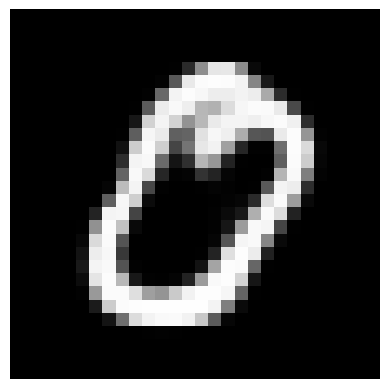}\\
      \includegraphics[scale=0.12]{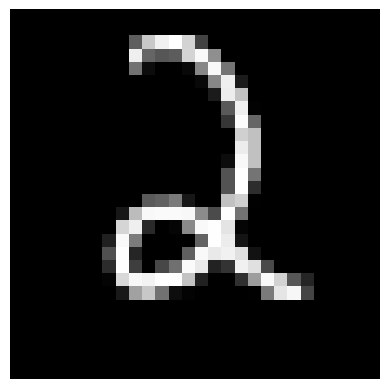} &\includegraphics[scale=0.12]{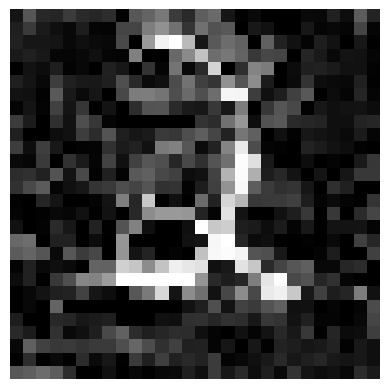} &\includegraphics[scale=0.12]{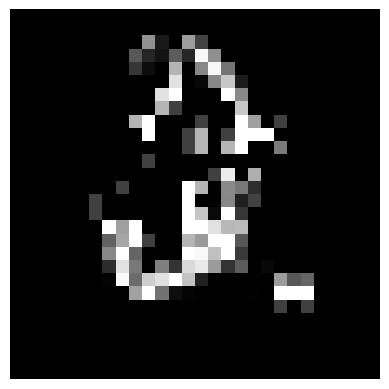} &\includegraphics[scale=0.12]{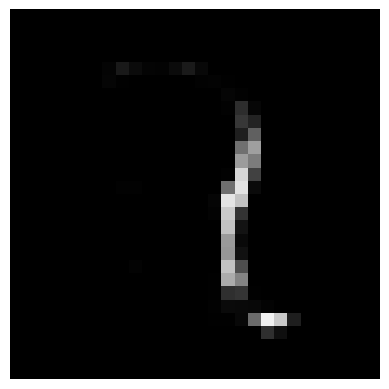}
      &\includegraphics[scale=0.12]{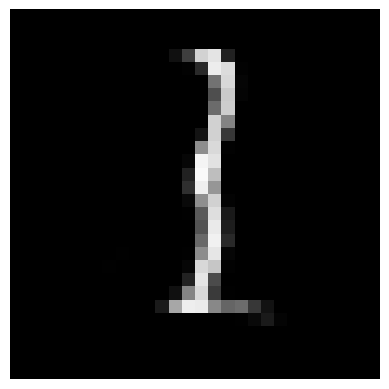}\\
      \includegraphics[scale=0.12]{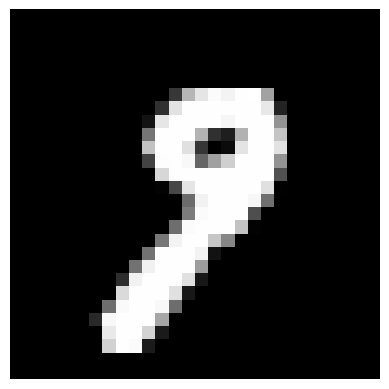} &\includegraphics[scale=0.12]{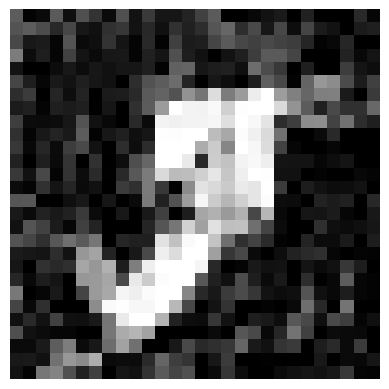} &\includegraphics[scale=0.12]{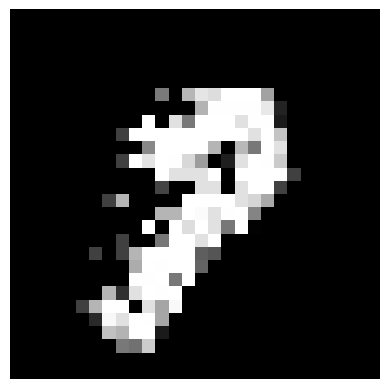} &\includegraphics[scale=0.12]{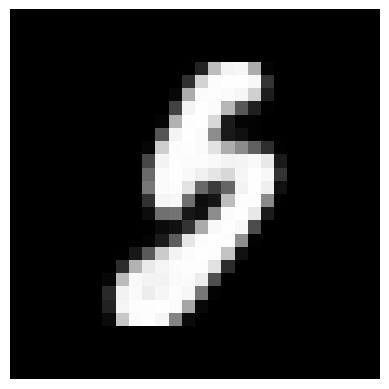}
      &\includegraphics[scale=0.12]{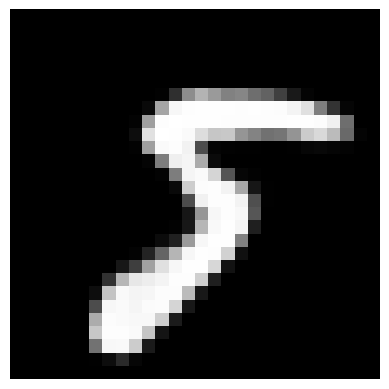}\\
      \\
      \hline
  \end{tabularx}
\end{center}
\label{table:mnist}
\end{table}

Table \ref{table:celeba} shows results on the CelebA dataset \citep{Liu15}. We used a $\beta$-VAE  \citep{Higgins16} with the same architecture proposed by \citet{Subramanian15}. The latent space itself is of dimension 128. We used, for both REVISE-E and Clarity, an ensemble of 20 models with adversarial training. There are two versions of REVISE-E: R-E (image) is using a penalty on the image space $d(\mathcal{G}_\psi(z'), X)$, and R-E (latent) is using a penalty on the latent space $d(z',z)$, similarly to Clarity. The classifiers were trained to predict the hair color, and as such, the goal of these explanations is to change it without modifying anything else. Overall, both methods offer similar results, but Clarity can be closer to the original image, unlike REVISE which can sometimes change the gender or the face of the person or add makeup. These results are mostly preliminary and might need to be confirmed on a VAE that performs better reconstruction with less blur.
\begin{table}[htbp]
\caption{Hair color counterfactual explanation comparison on the CelebA dataset. From top to bottom, the explanations are: Blond $\rightarrow$ Black, Black $\rightarrow$ Grey, Black $\rightarrow$ Blond. Columns represent the original image, the same image reconstructed by the VAE and the images obtained by the different counterfactual explanation methods.}
\begin{center}
  \begin{tabularx}{\linewidth}{*{5}{Xc}}
      \hline Original & VAE & R-E & R-E & Clarity \\
      & & (image) & (latent) & \\
      \hline \\
      \includegraphics[scale=0.5]{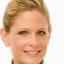} &\includegraphics[scale=0.5]{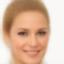} &\includegraphics[scale=0.5]{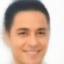}  &\includegraphics[scale=0.5]{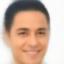} 
      &\includegraphics[scale=0.5]{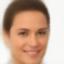}\\
      \includegraphics[scale=0.5]{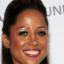} &\includegraphics[scale=0.5]{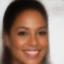} &\includegraphics[scale=0.5]{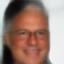}  &\includegraphics[scale=0.5]{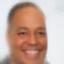} 
      &\includegraphics[scale=0.5]{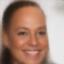}\\
      \includegraphics[scale=0.5]{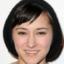} &\includegraphics[scale=0.5]{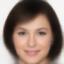} &\includegraphics[scale=0.5]{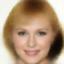}  &\includegraphics[scale=0.5]{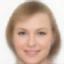} 
      &\includegraphics[scale=0.5]{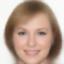}\\
      \\
      \hline
  \end{tabularx}
\end{center}
\label{table:celeba}
\end{table}

\section{DISCUSSION}
In this section, we will discuss the possible implications of the practical use of these findings, the current limitations of our method as well as potential directions  for further research. While counterfactual explanations are a useful tool to debug deep neural networks and to show modes of failure, they are often useless to end users if they do not make sense to them. Therefore, we hope that the results shown in this paper will help to put in place models that are designed to be understandable to any person. In addition, we believe that models interpretable ``by design'' might be a key to produce fairer models that do not discriminate in salient/sensible features irrelevant to the prediction task.

On the other hand, our approach is dependent on the presence of a generative model that, if not trained properly, could lead to undesirable side effects, such as introduction of possible biases in the generation process. For instance, a malicious user could train the generative model in a biased manner to create a classifier that is also biased and unfair by design. 
Another limitation of our method is the fact that the level of detail, and thus, the level of realism, of the produced counterfactuals rely on the quality of the generative model of the VAE, which is known to be inferior to other types of generative models such as GANs. Special care must therefore be taken in the training stage of the VAE.

This last remark leads us to a perspective: all of the techniques presented on this paper could be adapted to any generative model, such as GANs or Normalizing Flows \citep{Rezende15} or more sophisticated VAE architectures \citep{Larsen16, Dai19}. Several counterfactual explanation methods already use GANs \citep{Liu19, Lang21} and even diffusion models \citep{Sanchez22}. This opens several new and interesting directions of further research, namely, investigating the behavior of these techniques on such generative models, in order to overcome the limitations of VAEs in terms of image sharpness and level of detail.

\section{CONCLUSION}
In this paper we propose a method based on generative models and classifier ensembles, that improves the quality of counterfactual visual explanations. 
This method contributes to make models more interpretable by design, and capable of generating realistic and unambiguous counterfactual images  with minimal unnecessary changes. Moreover, such counterfactual images are an order of magnitude faster to compute. 
At a more fundamental level, we give insights, supported by experiments, into the reason why applying a gradient-based method to a classifier trained in the latent space of a VAE is more likely to produce quality counterfactual images in comparison to a classifier trained in the image space.

\subsubsection*{References}

		  
\begingroup
\renewcommand{\section}[2]{}%

\endgroup


\begin{thebibliography}{}
\setlength{\itemindent}{-\leftmargin}
\makeatletter\renewcommand{\@biblabel}[1]{}\makeatother

\bibitem[Boreiko et~al.(2022)]{Boreiko22}
V.~Boreiko, M.~Augustin, F.~Croce, P.~Berens, and M.~Hein.
\newblock Sparse Visual Counterfactual Explanations in Image Space.
\newblock \textit{arXiv preprint arXiv:2205.07972}, 2022.
    
\bibitem[Callahan and Shah(2017)]{Callahan17}
A.~Callahan, and N.~H.~Shah.
\newblock Chapter 19 - Machine Learning in Healthcare.
\newblock In A.~Sheikh, K.~M.~Cresswell, A.~Wright, and D.~W.~Bates,
\textit{Key Advances in Clinical Informatics}, pages 279--291, Academic Press, 2017.
    
\bibitem[Chang et~al.(2019)]{Chang19}
C.~H.~Chang, E.~Creager, A.~Goldenberg, and D.~Duvenaud.
\newblock Explaining Image Classifiers by Counterfactual Generation. 
\textit{Proceedings of International Conference on Learning Representations (ICLR)}, 2019.

\bibitem[Dai and Wipf(2019)]{Dai19}
B.~Dai, and D.~P.~Wipf.
\newblock Diagnosing and Enhancing VAE Models.
\textit{Proceedings of International Conference on Learning Representations (ICLR)},
2019.

\bibitem[Depeweg et~al.(2018)]{Depeweg18}
S.~Depeweg, J.~M.~Hern{\'{a}}ndez{-}Lobato, F.~Doshi{-}Velez, and S.~Udluft.
\newblock Decomposition of uncertainty in bayesian deep learning for efficient and risk-sensitive learning. 
\newblock In J.~G.,~Dy, A.~Krause,
\textit{Proceedings of International Conference on Machine Learning (ICML)},
pages 1192--1201, 2018.

\bibitem[Freiesleben(2022)]{Freiesleben22}
T.~Freiesleben.
\newblock The Intriguing Relation Between Counterfactual Explanations and Adversarial Examples.
\newblock In \textit{Minds \& Machines} 32, pages 77--109 (2022).
    
       
\bibitem[Goodfellow et~al.(2014)]{Goodfellow14} I.~J.~Goodfellow, J.~Pouget-Abadie, M.~Mirza, B.~Xu, D.~Warde-Farley, S.~Ozair, A.~C.~Courville, and Y.~Bengio.
\newblock Generative Adversarial Nets.
\newblock In Z.~Ghahramani, M.~Welling, C.~Cortes, N.~D.~Lawrence, and K.~Q.~Weinberger,
\textit{Proceedings of the International Conference on Neural Information Processing Systems (NIPS)},
pages 2672--2680, 2014.

\bibitem[Goodfellow et~al.(2015)]{Goodfellow15}
I.~J.~Goodfellow, J.~Shlens, and C.~Szegedy.
\newblock Explaining and Harnessing Adversarial Examples.
\newblock In Y.~Bengio, Y.~LeCun,
\textit{Proceedings of International Conference on Learning Representations (ICLR)}, 2015.
 
\bibitem[Goyal et~al.(2019)]{Goyal20}
Y.~Goyal, Z.~Wu, J.~Ernst, D.~Batra, D.~Parikh, and S.~Lee.
\newblock Counterfactual Visual Explanations.
\newblock In K.~Chaudhuri, and R.~Salakhutdinov,
\textit{Proceedings of International Conference on Machine Learning (ICML)},
pages 2376--2384, 2019.
    
\bibitem[He et~al.(2015)]{He15}
K.~He, X.~Zhang, S.~Ren, and J.~Sun.
\newblock Delving Deep into Rectifiers: Surpassing Human-Level Performance on ImageNet Classification.
\textit{Proceedings of the International Conference on Computer Vision (ICCV)},
Pages 1026--1034, 2015.

\bibitem[Higgins et~al.(2016)]{Higgins16}
I.~Higgins, L.~Matthey, A.~Pal, C.~P.~Burgess, X.~Glorot, M.~M.~Botvinick, S.~Mohamed, and A.~Lercher.
\newblock beta-VAE: Learning Basic Visual Concepts with a Constrained Variational Framework.
\textit{Proceedings of International Conference on Learning Representations (ICLR)}, 2017.

\bibitem[H{\"{o}}ltgen et~al.(2021)]{Holtgen21}
B.~H{\"{o}}ltgen, L.~Schut, J.~M.~Brauner, and Y.~Gal.
\newblock DeDUCE: Generating Counterfactual Explanations Efficiently
\newblock \textit{arXiv preprint arXiv:2111.15639}, 2021.
    
\bibitem[Ioffe and Szegedy.(2015)]{Ioffe15}
S.~Ioffe, and C.~Szegedy.
\newblock Batch Normalization: Accelerating Deep Network Training by Reducing Internal Covariate Shift.
\newblock In F.~R.~Bach, and D.~M.~Blei,
\textit{Proceedings of International Conference on Machine Learning (ICML)},
pages 448--456, 2015.

\bibitem[Joshi et al.(2019)]{Joshi19}
S.~Joshi, O.~Koyejo, W.~Vijitbenjaronk, B.~Kim, and J.~Ghosh.
\newblock Towards Realistic Individual Recourse and Actionable Explanationsin Black-Box Decision Making Systems
\newblock \textit{arXiv preprint arXiv:1907.09615}, 2019.

\bibitem[Keane et al.(2021)]{Keane21}
M.~T.~Keane, E.~M.~Kenny, E.~Delaney, and B.~Smyth.
\newblock If Only We Had Better Counterfactual Explanations: Five Key Deficits
to Rectify in the Evaluation of Counterfactual XAI Techniques.
\newblock In Z.-H. Zhou,
\textit{Proceedings of the International Joint Conference on Artificial Intelligence (IJCAI)},
pages 4466--4474, 2021.

\bibitem[Kendall and Gal(2017)]{Kendall17}
A.~Kendall, and Y.~Gal.
\newblock What Uncertainties Do We Need in Bayesian Deep Learning for Computer Vision ?
\newblock In I.~Guyon, U.~von Luxburg, S.~Bengio, H.~M.~Wallach, R.~Fergus, S.~V.~N.~Vishwanathan, and R.~Garnett,
\textit{Proceedings of the International Conference on Neural Information Processing Systems (NIPS)},
pages 5574--5584, 2017.

\bibitem[Kingma and Ba(2015)]{Kingma15}
D.~P.~Kingma, and J.~Ba.
\newblock Adam: A Method for Stochastic Optimization.
\newblock In Y.~Bengio, and Y.~LeCun,
\textit{Proceedings of International Conference on Learning Representations (ICLR)}, 2015.
    
\bibitem[Kingma and Welling(2014)]{Kingma14}
D.~P.~Kingma, and M.~Welling.
\newblock Auto-Encoding Variational Bayes.
\newblock In Y.~Bengio, and Y.~LeCun,
\textit{Proceedings of International Conference on Learning Representations (ICLR)}, 2014.
    
\bibitem[Lakshminarayanan et~al.(2017)]{Laksh17}
B.~Lakshminarayanan, A.~Pritzel, and C.~Blundell.
\newblock Simple and scalable predictive uncertainty estimation using deep ensembles.
\newblock In I.~Guyon, U.~von Luxburg, S.~Bengio, H.~M.~Wallach, R.~Fergus, S.~V.~N.~Vishwanathan, and R.~Garnett,
\textit{Proceedings of the International Conference on Neural Information Processing Systems (NIPS)},
pages 6402--6413, 2017. 

\bibitem[Lang et~al.(2021)]{Lang21}
O.~Lang, Y.~Gandelsman, M.~Yarom, Y.~Wald, G.~Elidan, A.~Hassidim, W.~T.~Freeman, P.~Isola, A.~Globerson, M.~Irani, and I.~Mosseri.
\newblock Explaining in Style: Training a GAN to Explain a Classifier in StyleSpace.
\textit{Proceedings of the International Conference on Computer Vision (ICCV)},
pages 673--682, 2021.
    
\bibitem[Larsen et~al.(2016)]{Larsen16}
A.~B.~L.~Larsen, S.~K.~Sønderby, J.~Larochelle, and O.~Winther.
\newblock Autoencoding Beyond Pixels Using a Learned Similarity Metric.
\newblock In: M.-F.~Balcan, and K.~Q.~Weinberger,
\textit{Proceedings of the International Conference on Machine Learning (ICML)},
pages 1558--1566, 2016.


\bibitem[LeCun et~al.(2010)]{mnist}
Y.~LeCun, C.~Cortes, and C.~Burges.
\newblock MNIST handwritten digit database.
\newblock ATT Labs (http://yann.lecun.com/exdb/mnist). 2010.
    
\bibitem[Liu et~al.(2015)]{Liu15}
Z.~Liu, P.~Luo, X.~Wang, and X.~Tang.
\newblock Deep Learning Face Attributes in the Wild.
\textit{Proceedings of the International Conference on Computer Vision (ICCV)}, 2015.

\bibitem[Liu et~al.(2019)]{Liu19}
S.~Liu, B.~Kailkhura, D.~Loveland and Y.~Han.
\newblock Generative Counterfactual Introspection for Explainable Deep Learning.
\textit{The IEEE Global Conference on Signal and Information Processing (GlobalSIP)}, pages 1--5, 2019.

\bibitem[Papernot et~al.(2016)]{Papernot16}
N.~Papernot, P.~McDaniel, S.~Jha, M.~Frederikson, Z.~B~.Celik, and A.~Swani.
\newblock The Limitations of Deep Learning in Adversarial Settings.
\textit{Proceedings of the IEEE European Symposium on Security and Privacy},
pages 372--387, 2016.

\bibitem[Rezende and Blei(2015)]{Rezende15}
D.~Rezende, and S.~Mohamed.
\newblock Variational Inference with Normalizing Flows.
\textit{Proceedings of the International Conference on Machine Learning} (ICML),
page 1530--1538, 2015.
    
\bibitem[Sanchez and Tsaftaris(2022)]{Sanchez22}
P.~Sanchez, and S.~A.~Tsaftaris.
\newblock Diffusion Causal Models for Counterfactual Estimation.
\textit{First Conference on Causal Learning and Reasoning}, 2022.

\bibitem[Sartor and Lagioia(2020)]{Sartor20}
G.~Sartor, and F.~Lagioia.
\newblock The Impact of the General Data Protection Regulation (GDPR) on Artificial Intelligence
\newblock \textit{European Parliamentary Research Service}, pages 1--100, 2020.

\bibitem[Schut et~al.(2021)]{Schut21}
L.~Schut, O.~Key, R.~McGrath, L.~Costabello, B.~Sacaleanu, M.~Corcoran, and Y.~Gal.
\newblock Generating Interpretable Counterfactual Explanations By Implicit Minimisation of Epistemic and Aleatoric Uncertainties.
\newblock In A.~Banerjee, and K.~Fukumizu,
\textit{Proceedings of the International Conference on Artificial Intelligence and Statistics (AISTATS)}, pages 1756--1764, 2021.
    
\bibitem[Singla(2020)]{Singla20}
S.~Singla.
\newblock Machine Learning to Predict Credit Risk in Lending Industry.
\newblock https://www.aitimejournal.com/@saurav.singla/machine-learning-to-predict-credit-risk-in-lending-industry
    
\bibitem[Smith and Gal(2018)]{Smith18} L.~Smith, and Y.~Gal.
\newblock Understanding Measures of Uncertainty for Adversarial Example Detection.
\newblock In A.~Globerson, R.~Silva,
\textit{Proceedings of the International Conference on Uncertainty in Artificial Intelligence (UAI)}, pages 560--596, 2018.
    
\bibitem[Subramanian(2020)]{Subramanian15}
A.~K.~Subramanian.
\newblock PyTorch-VAE (2020).
\newblock \url{https://github.com/AntixK/PyTorch-VAE}

\bibitem[Szegedy et~al.(2014)]{Szegedy14}
C.~Szegedy, W.~Zaremba, I.~Sutskever, J.~Bruna, D.~Erhan, I.~J.~Goodfellow, and R.~Fergus.
\newblock Intriguing properties of neural networks.
\textit{Proceedings of International Conference on Learning Representations (ICLR)}, 2014.
    
\bibitem[Van Looveren and Klaise(2021)]{Looveren21}
A.~Van Looveren, and J.~Klaise.
\newblock Interpretable Counterfactual Explanations Guided by Prototypes
\newblock In N.~Oliver, F.~P{\'{e}}rez{-}Cruz, S.~Kramer, J.~Read, and J.~Antonio Lozano,
\textit{Proceedings of Machine Learning and Knowledge Discovery in Databases (ECML-PKDD)}.
Springer LNCS 12976, pages 650--655, 2021. 
    
\bibitem[Watcher et~al.(2017)]{Watcher17}
S.~Watcher, B.~D.~Mittelstadt, and C.~Russell (2017).
\newblock Counterfactual Explanations without Opening the Black Box: Automated Decisions and the GDPR.
\newblock \textit{Harvard Journal of Law \& Technology}, 31:841–887, 2018.

\end{thebibliography}
\end{document}